%% file: main.tex
\documentclass[sn-nature]{sn-jnl}
\usepackage{graphicx}%
\usepackage{multirow}%
\usepackage{amsmath,amssymb,amsfonts}%
\usepackage{amsthm}%
\usepackage{mathrsfs}%
\usepackage[title]{appendix}%
\usepackage{xcolor}%
\usepackage{textcomp}%
\usepackage{manyfoot}%
\usepackage{booktabs}%
\usepackage{algorithm}%
\usepackage{algorithmicx}%
\usepackage{algpseudocode}%
\usepackage{listings}%

\usepackage{dcolumn}
\usepackage{bm}
\usepackage{xcolor}
\usepackage{subfigure}
\usepackage[utf8]{inputenc}
\usepackage[T1]{fontenc}
\usepackage{float}

\usepackage{mathptmx}
\usepackage{etoolbox}
\usepackage{caption}
\usepackage{multirow}

\input{commands}

\begin{document}
\input{abstract}

\title{PolyGET: Accelerating Polymer Simulations by Accurate and Generalizable Forcefield with
Equivariant Transformer}

\author[1]{\fnm{Rui} \sur{Feng}}\email{rfeng@gmail.com}
\author[1]{\fnm{Huan} \sur{Tran}}\email{huan.tran@mse.gatech.edu}
\author[1]{\fnm{Aubrey} \sur{Toland}}\email{atoland@gmail.com}
\author[1]{\fnm{Binghong} \sur{Chen}}\email{bchen@gmail.com}
\author[2]{\fnm{Qi} \sur{Zhu}}\email{qzhu@gmail.com}
\author*[1]{\fnm{Rampi} \sur{Ramprasad}}\email{rramprasad@gmail.com}
\author*[1]{\fnm{Chao} \sur{Zhang}}\email{czhang@gmail.com}


\affil*[1]{\orgdiv{Georgia Institute of Technology}, \orgname{Georgia Institute of Technology}, \orgaddress{\street{225 North Avenue NW}, \city{Atlanta}, \postcode{30332}, \state{GA}, \country{USA}}}

\affil[2]{\orgdiv{University of Illinois Urbana-Champaign}, \orgname{University of Illinois Urbana-Champaign}, \orgaddress{\street{601 East John Street}, \city{Champaign}, \postcode{61820}, \state{IL}, \country{USA}}}


\maketitle

\input{intro_new}
\input{short_versions/method}

\input{data}

\input{short_versions/experiment}

\input{related}
\input{conclusion.tex}

\bibliography{references}

\end{document}

%% file: commands.tex
\newcommand{\ours}{PolyGET}
\newcommand{\ourdata}{Poly24}

\newcommand{\hpolye}[1]{}

\newcommand{\RR}{\mathbb{R}}

\newcommand{\expt}{\mathbb{E}}
\newcommand{\thermo}{\mathcal{T}}

\newcommand{\DFT}{\mathrm{DFT}}
\newcommand{\ML}{\mathrm{ML}}

\newcommand{\hide}[1]{}

%% file: abstract.tex
\abstract{
 Polymer simulation with both accuracy and efficiency is a challenging task.
 Machine learning (ML) forcefields  have been developed to achieve both the accuracy of ab initio methods  and the efficiency of empirical force fields. 
   However, existing ML force fields are usually limited to single-molecule settings, and their simulations are not robust enough. In this paper, we present PolyGET, a new framework for Polymer Forcefields with Generalizable Equivariant Transformers. PolyGET is designed to capture complex quantum interactions between atoms and generalize across various polymer families, using a deep learning model called Equivariant Transformers. We propose a new training paradigm that focuses exclusively on optimizing forces, which is different from existing methods that jointly optimize forces and energy. This simple force-centric objective function avoids competing objectives between energy and forces, thereby allowing for learning a unified forcefield ML model over  different polymer families. We evaluated PolyGET on a large-scale dataset of 24 distinct polymer types and demonstrated state-of-the-art performance in force accuracy and robust MD simulations. Furthermore, PolyGET can simulate large polymers with high fidelity to the reference ab initio DFT method while being able to generalize to unseen polymers. }

%% file: intro_new.tex
\section{Introduction}
\label{sect:intro}
Efficient and accurate forcefield computation is essential for polymer
simulation, but it remains a challenging task. Although ab initio methods
offer excellent accuracy, they are computationally slow. For example,
Density Functional Theory (DFT)\cite{DFT1,DFT2} has a time complexity of
\(N_e^3\), where \(N_e\) is the number of electrons, rendering it too
expensive for large polymers or long-horizon simulations. In contrast,
empirical forcefields are more computationally efficient, as they simplify
chemical interactions between atoms to a sum over bonded and unbonded
pairwise terms\cite{unke2020high,gonzalez2011force}. This
approach reduces the time complexity to \(N_a^2\), where \(N_a\) is the
number of atoms, allowing for simulating larger molecular systems. However, their accuracy is limited due to the simplification of
chemical interactions, which results in suboptimal molecular dynamics (MD)
simulation performance and chemical insights~\cite{unke2021machine}.

Machine learning (ML) forcefields combine ab initio accuracy with empirical forcefield computational efficiency. Early methods, such as kernel regression~\cite{chmiela2017machine,chmiela2018towards,chmiela2022accurate} and feedforward neural networks~\cite{svozil1997introduction}, rely on hand-crafted features, limiting flexibility and expressivity for complex molecular structures and interactions~\cite{unke2021machine}. Recent Equivariant Graph Neural Networks (EGNNs)\cite{tholke2022torchmd,fuchs2020se,batzner20223,satorras2021n} enable automatic feature learning, achieving state-of-the-art accuracy in predicting energy and forces. However, existing EGNNs face two main limitations: (1) They are trained in a single-molecule setting, necessitating separate models for different molecules and lacking generalization to unseen molecules. In Section\ref{sec:exp:multimolecule}, we will demonstrate that adapting EGNNs to the multi-molecule paradigm is nontrivial due to conflicting objectives in optimizing energy and force. (2) They are not robust enough for simulations~\cite{fu2022forces}, despite achieving high accuracy in static forcefield prediction. The single-molecule training paradigm can lead to overfitting to individual molecules, compromising the model's ability to perform robustly in dynamic simulations.
\begin{figure}[h]
  \centering
  \includegraphics[width=0.99\linewidth]{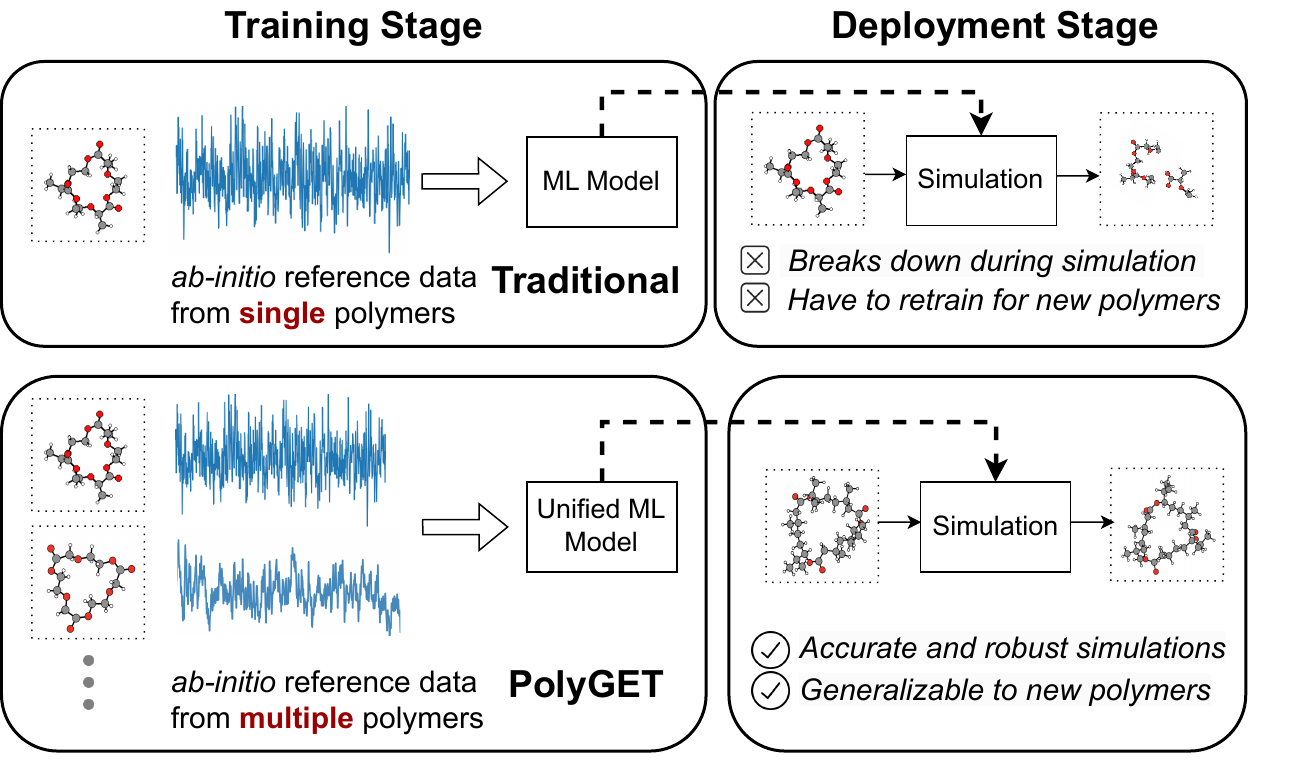}
  \caption{PolyGET learns a single forcefield model across various polymer families by capturing quantum mechanical interactions. The unified model captures generalizable knowledge from ab-initio reference calculations. PolyGET enables accurate and reliable MD simulations while being able to generalize to large, unseen polymers.}
  \label{fig:paradigm}
\end{figure}

We introduce a new ML forcefield approach for polymer simulation, named \emph{\textbf{Poly}mer Forcefields with \textbf{G}eneralizable \textbf{E}quivariant \textbf{T}ransformers} (\ours). 
As depicted in Figure \ref{fig:paradigm}, 
\ours~has two key advantages compared to existing EGNN models: (1) \emph{Multi-molecule training and generalization}: The model trains over a diverse set of polymer families and can generalize to previously unseen and larger polymers from smaller ones. (2) \emph{Accurate and robust simulations}: The model can produce reliable and precise forces during molecular dynamics (MD) simulations. \ours~achieves these advantages through two novel designs:

First, we train a unified multi-molecule model using Equivariant Transformer \cite{tholke2022torchmd} as the backbone model. As a variant of the Transformer model \cite{vaswani2017attention} based on the attention mechanism, the Equivariant Transformer facilitates learning interactions between atoms. By incorporating multiple layers of attention blocks, the model effectively captures complex quantum interactions. Furthermore, the Equivariant Transformer learns latent atomic features that are roto-translational equivariant to atomic positions, ensuring the invariance of the learned energy, which is necessary for energy conservation. Its expressivity and capacity make it a suitable backbone for \ours.

Second, we propose a novel training paradigm that facilitates multi-molecule training with a unified ML model. While existing deep learning approaches \cite{tholke2022torchmd,schutt2018schnet} jointly optimize potential energy and forces, our training paradigm focuses solely on the optimization of forces, which we have found to be generalizable across multiple polymer families. With our training paradigm, the underlying ML model acquires robust and transferable knowledge about quantum mechanical interactions between atoms, enabling both accurate and robust MD simulations and the capacity to generalize to unseen and larger polymers with similar quantum interactions. Although our model concentrates on force optimization, it can also output energy that is linearly correlated with ground truth energy, allowing the model to estimate ground truth energy with numerical integration.

We have trained and evaluated the \ours~model on \ourdata, a novel benchmark comprising 24 polymer types and 6,552,624 conformations from four categories: cycloalkanes, lactones, ethers, and others. \ours~demonstrates state-of-the-art force accuracy and generalizes to previously unseen large polymers. Furthermore, we have assessed the capacity of \ours~to perform accurate MD simulations. We found that even for large 15-loop polymers not present in the training data, \ours~can simulate polymers with high fidelity to the reference DFT data.

In summary, we introduce a novel \ours~for polymer forcefields, achieving: (1) a unified Equivariant Transformer model that can transfer across multiple polymer families for forcefield and energy prediction, (2) accurate and robust MD simulation for large unseen polymers, and (3) state-of-the-art performance in force prediction and MD simulation on a large benchmark \ourdata. This paradigm can be applied to various deep learning models, enabling the development of general ML forcefields and robust MD simulation for a broader range of chemical families.

%% file: short_versions/method.tex
\section{The \ours ~Model}

\subsection{Force-centric Multi-molecule Optimization}
\label{sec:method:paradigm}

Let 
\(x\)  denote a molecule consisting of 
\(n_x\)
 atoms. The conformations of these atoms are represented by 
\(\mathbf{r}_x\in\RR^{n_x\times 3}\)
 and their atom types are denoted by 
\(\mathbf{z}_x\in\RR^{n_x}\). The potential energy, 
\(E\), is a function of the conformations and atom types. The forces exerted on the atoms are derived by
\(F=-\frac{\partial{E}}{\partial\mathbf{r}}\).
In molecular dynamics (MD) simulations, molecular conformations are influenced by the forces applied and the simulation thermostat.
Define 
\(\mathbb{T}^{\DFT}(x,\mathbf{r}_0, \thermo)\)
 as the distribution of conformations in a density functional theory (DFT)-based MD simulation for a specific thermostat 
$\thermo$ and initial conformation 
\(\mathbf{x}_0\).
Machine learning (ML) force fields aim to learn a model, 
\(\Phi_\theta\), parameterized by $\theta$,
which approximates DFT-generated reference data. Existing ML force field models optimize model parameters 
 $\theta$
 by minimizing the following objective:

\begin{equation}
\theta = \arg\min \expt^{\DFT} \left[ \Vert E^{\ML} - E^{\DFT}\Vert^2_2 + \Vert F^{\ML} - F^{\DFT} \Vert^2_2\right]. \label{eqn:mainstrem_optim}
\end{equation}

This optimization process is typically conducted in single-molecule settings by existing EGNN models, which learn a separate model for each molecule. As discussed in Section \ref{sect:intro}
 and will be demonstrated in Section \ref{sect:result}, models trained in a single-molecule setting not only fail to generalize to unseen molecules but also perform poorly in simulations due to their tendency to overfit the single molecules.
Although it is natural to question whether a unified model can be trained to leverage DFT reference data for multiple molecules, this task is nontrivial. The challenge of training a unified model for multiple molecules can be understood by analyzing the per-atom potential energy and force distributions. As illustrated in Figure~\ref{fig:prelim:dist}, although forces follow similar distributions for different types of polymers, the per-atom potential energy does not. We hypothesize that the mismatch between potential energy and force distributions is the primary cause of the optimization bottleneck and the reason for the preference for single-molecule settings in existing methods. In Section~\ref{sec:exp:multimolecule}, we demonstrate that when extending existing EGNN methods for multi-molecule training---\emph{i.e.}, training a unified ML force field for various types of molecules---their performance can deteriorate instead of improving.

\begin{figure}[h]
\centering \includegraphics[width=0.9\linewidth]{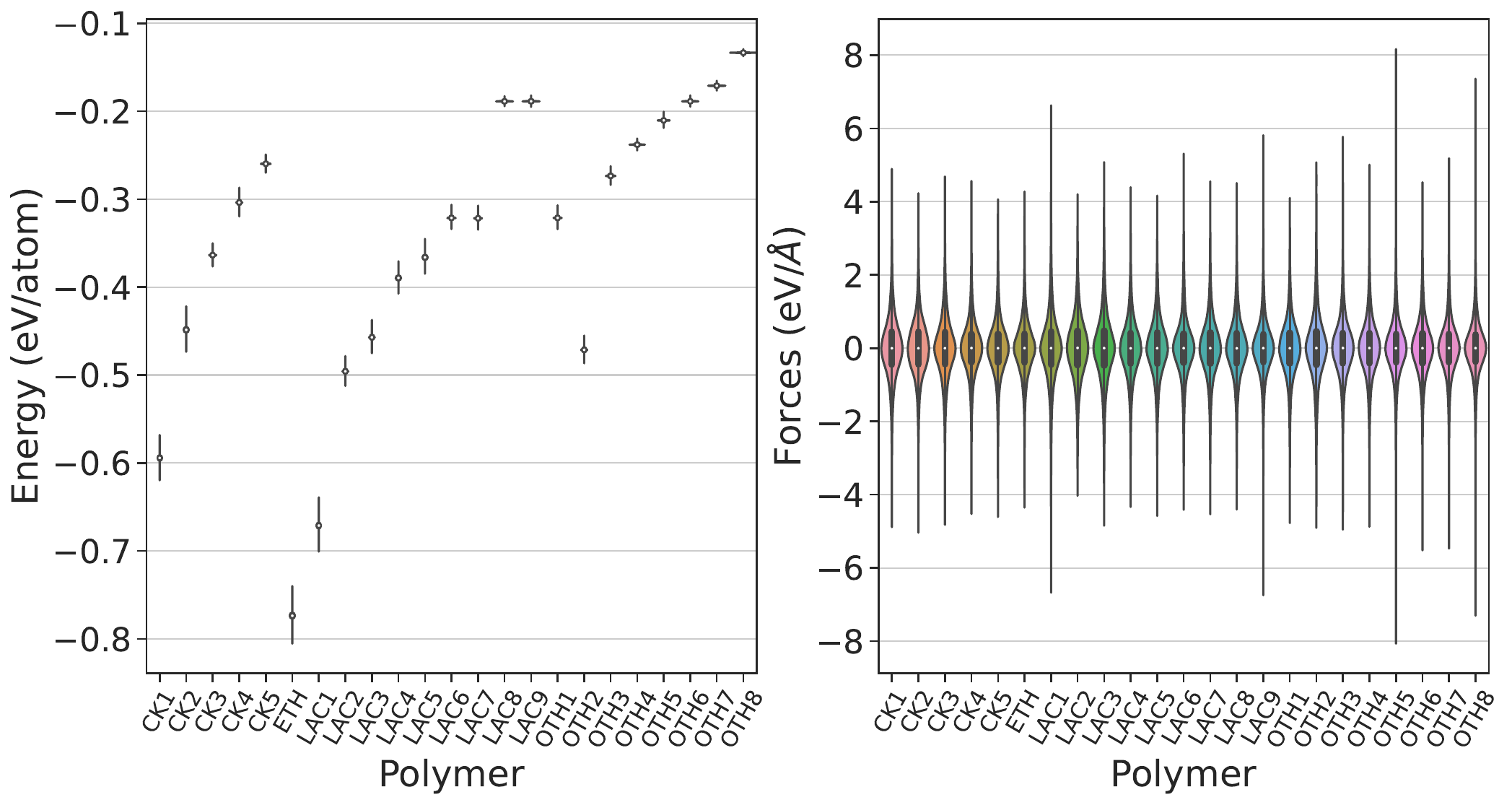}\label{fig:prelim:dist}
\caption{The distribution of per-atom potential energies and forces across different polymers. Forces exhibit similar distributions for various types of polymers, whereas the per-atom potential energy does not.} 
\label{fig:prelim:dist}
\end{figure}

In our approach, we address this challenge and learn generalizable force fields from multiple molecules by proposing a new learning objective:

\begin{equation}
\expt_{x, \mathbf{
r}0} \left[\expt_{\mathbf{r}\sim \mathbb{T}^{\DFT}(x, \mathbf{r}0, \thermo)}\left[ \left\Vert - \nabla\Phi\theta(\mathbf{r}, \mathbf{z}) - F(\mathbf{r}, \mathbf{z}) \right\Vert^2_2 \right] \right].
\label{eqn:multi_mole_obj}
\end{equation}

In this objective, we minimize the differences between the predicted forces
\(-\nabla \Phi_\theta (\mathbf{r}, \mathbf{z})\) and the ground-truth forces \(F(\mathbf{r},\mathbf{z})\).
 The loss is optimized across multiple DFT trajectories for different types of polymers.

The primary distinction between our approach and existing methods lies in
the exclusive focus on forces, rather than jointly optimizing the potential
energy and forces. This design choice is grounded in two key rationales: 1)
forces, being a local atomic attribute, exhibit greater transferability
across diverse molecules~\cite{ramprasad2017machine,botu2017machine}, and
2) forces demonstrate similar distributions among different molecules, as
evidenced in Figure \ref{fig:prelim:dist}. Conversely, the potential energy, a molecular-level
attribute, displays considerable variation in its distributions—even after
normalization—due to the intricate quantum mechanical processes involved.
By focusing on forces during training, our model can efficiently transfer information across DFT-generated data without being hindered by competing objectives between energy and force. In order to perform well across a wide array of molecules and conformation distributions, the unified model must effectively capture generic atomic interactions, rather than overfitting to a single molecule. As a result, this comprehensive training on various molecules bolsters our model's capacity to generalize to unseen molecules and conduct more robust simulations.

\begin{figure}[h!]
    \centering
    \includegraphics[width=0.99\linewidth]{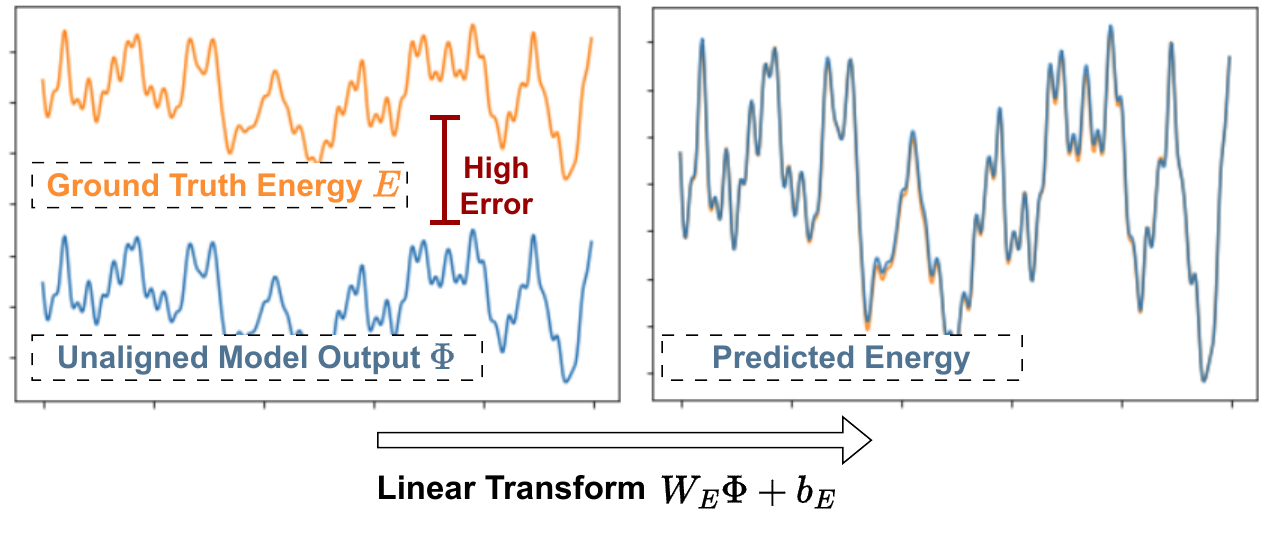}
    \caption{The model output is not directly optimized to be aligned with the ground truth potential energy but can approximate the latter with a linear transform.}
    \label{fig:model:energy_prediction}
\end{figure}

\subsection{Energy Prediction}

The objective function~\eqref{eqn:multi_mole_obj} provides a supervision signal to the model only on the forces.
In order to predict the potential energy for molecules during simulation, we observe that by training exclusively on forces with~\eqref{eqn:multi_mole_obj}, the model output \(\Phi_\theta\) is linearly correlated with the ground-truth potential energy \(E\) (Figure~\ref{fig:model:energy_prediction}).
We thus call the output energy from \(\Phi_\theta\) as \emph{uncalibrated potential energy}.
To predict  the true potential energy \(E\) from \(\Phi_\theta\),
we can estimate \(E_t\) with \({E}_t=E_0 + \int_0^t \frac{\partial \Phi_\theta}{\partial \mathbf{r}}(\mathbf{r}_s, \mathbf{z})ds\) for a fixed time window \(t\in [0, T]\), where \(E_0\) is the initial energy of at the start of the simulation. The integral can be approximated with
numeric Taylor expansion:
\[
  \overline{E}_t = \overline{E}_{t-1} - \left\langle \frac{\partial{\Phi}}{\partial{\mathbf{r}}}  (\mathbf{r}_{t-1}), \mathbf{r}_t - \mathbf{r}_{t-1} \right\rangle.
\]
Then given uniformly sampled \(0=t_0<\ldots<t_{m}=T\), we can estimate the potential energy function with
\begin{eqnarray}
  & E^*(\mathbf{r}, \mathbf{z}) & =  \mathbf{W}_E \Phi_\theta(\mathbf{r}, \mathbf{z}) + \mathbf{b}_E, \nonumber \\
  \mathrm{where}\  & (\mathbf{W}_E, \mathbf{b}_E) & =  \arg\min \sum_{i=0}^m \Vert (\mathbf{W}_E \Phi_\theta + \mathbf{b}_E) - \overline{E}_i\Vert.
\end{eqnarray}

\begin{figure}[h!]
    \centering
    \includegraphics[width=0.99\linewidth]{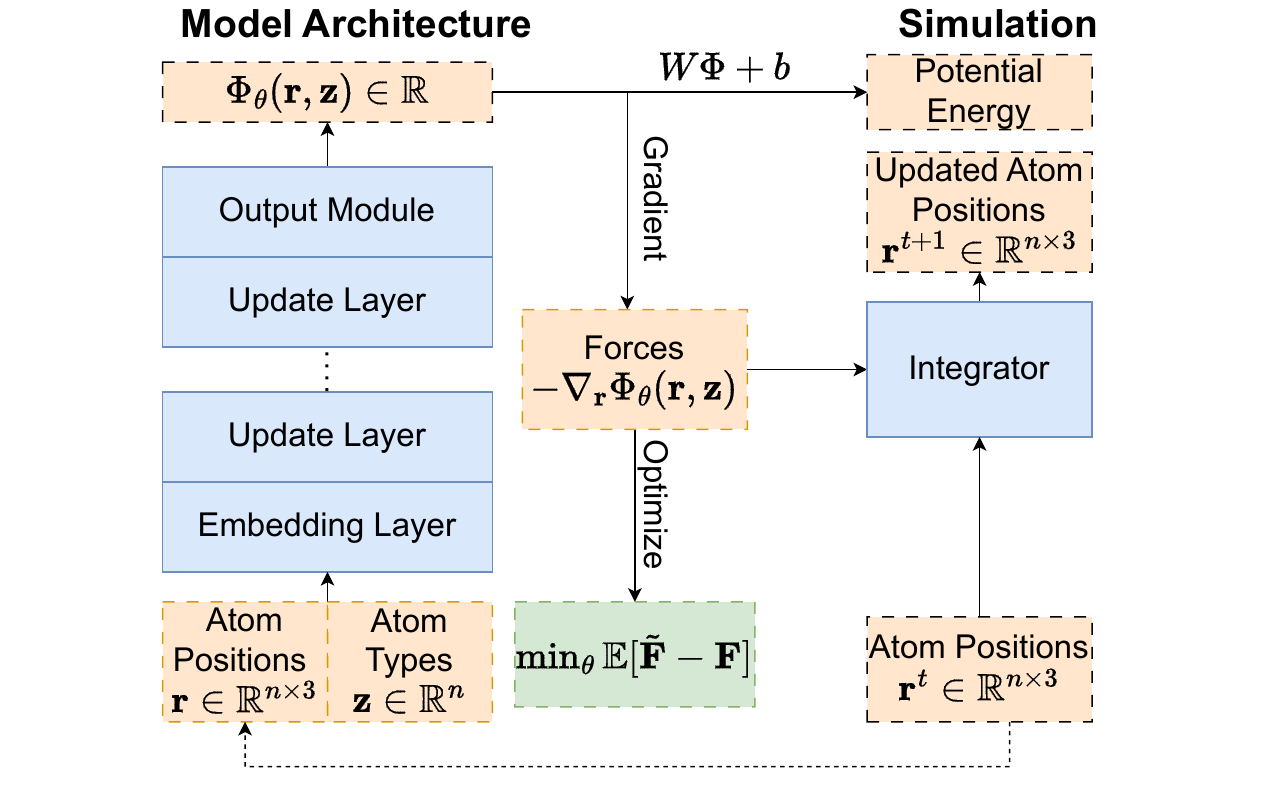}
    \caption{ Equivariant Transformer learns to predict a conservative vector field optimized to be atomic forces by taking the negative gradient of an invariant scalar output. The atomic forces can be used for MD simulations with an integrator.}
    \label{fig:model:archi}
\end{figure}

\section{Model architecture: Equivariant Transformer}
\label{sec:transformer}

In conjunction with our new learning objective, we use Equivariant
Transformer~\cite{tholke2022torchmd} as our backbone model for \ours.
Equivariant Transformer uses the powerful \emph{self-attention
  mechanism}~\cite{vaswani2017attention} and the Transformer architecture
for molecular data. The self-attention mechanism allows the model to
capture interatomic interactions and the attraction-repulsion dynamics.
Furthermore, Equivariant Transformer builds roto-translational equivariant
to atomic positions --- this not only allows it to learn in a more
data-efficient way, but also ensures the invariance of the learned energy
for the energy conservation law.

As shown in Figure \ref{fig:model:archi}, the Equivariant Transformer architecture operates on molecular data in the form of graphs and consists of three main blocks: an embedding layer, an updated layer, and an output network.

\paragraph*{Embedding Layer}
The first component is an atom embedding layer, which transforms atoms to vector representations that encode the quantum mechanical information about the atoms and their surroundings. The atom embedding is further split to \emph{scalar} embeddings \(\mathbf{h}_0\in\RR^D\) and \emph{vector} embeddings \(\mathbf{v}_0\in\RR^{3\times D}\), where \(D\) is the embedding size.
The scalar embedding is derived by integrating both the intrinsic vector, which captures information specific to the atoms themselves, and the neighborhood vector, which encapsulates the interactions within the atomic neighborhoods. The intrinsic vector, denoted as \(\mathbf{x}^\mathrm{int}\), is acquired by associating each atom type with a corresponding dense feature vector in \(\RR^D\). To compute the neighborhood embedding \(\mathbf{x}^\mathrm{nbr}\), we initially define the distance filter using a radial basis function. Subsequently, a continuous-filter convolution technique is employed to aggregate information from the surrounding atoms within a given neighborhood.
The atom embedding \(\mathbf{x}\) is obtained by the concatenation of intrinsic and neighborhood embeddings.
The vector embedding \(\mathbf{v}\) is initialized as zero vectors in \(\RR^{3\times D}\).

\paragraph*{Update Layers}
The update layers defines the incremental change of the \emph{scalar embeddings} \(\mathbf{x}\) and \emph{vector embeddings} \(\mathbf{v}\) in each network layer. For each atom \(i\), the update layers are defined as follows:
\begin{equation}
  \begin{aligned}
    \nabla \mathbf{x}_i & = f^{x}(\mathbf{x}_i) + \sum_{j} \phi^x(\mathbf{x}_i, \mathbf{x}_j) + \psi^{xv}(\mathbf{v}_i) \\
    \nabla \mathbf{v}_i & = f^{v}(\mathbf{v}_i) + \sum_{j} \frac{\mathbf{r}_j - \mathbf{r}_i}{\Vert \mathbf{r}_j - \mathbf{r}_i\Vert} \phi^{v}(\mathbf{x}_i, \mathbf{x}_j) + \psi^{vx}(\mathbf{x}_i).
  \end{aligned}
\end{equation}
For each atom \(i\), the attention mechanism \(\phi^x\) and \(\phi^v\) will model its interatomic potentials to all other atoms. Based on the embeddings of two atoms and their relative distance vectors, \(\phi^x\) and \(\phi^v\) define the interaction between the two atoms.
\(f^x\) and \(f^v\) are the residual connections~\cite{he2016identity} that prevent the network to suffer from performance degradation.
\(\psi^{xv}\) and \(\psi^{vx}\) define the information exchange between \(\mathbf{x}\) and \(\mathbf{v}\) in each layer. \(\psi^{xv}\) is used to update the vector embedding \(\mathbf{v}\) from the scalar embedding \(\mathbf{x}\), and \(\psi^{vx}\) is used to update the scalar embedding \(\mathbf{x}\) from the vector embedding \(\mathbf{v}\).

\paragraph*{Output Network}
The output network computes scalar atom-wise predictions using gated equivariant blocks, which is a modified version of the gated residual network~\cite{tholke2022torchmd}. The gated equivariant blocks aggregate the atomic embeddings into a single molecular prediction. This prediction can be matched with a scalar target variable or differentiated against atomic coordinates to provide force predictions.

\hide{
\begin{figure*}[tp]
  \centering
  \subfigure[Model Architecture.]{
    \includegraphics[width=0.60\linewidth]{figs/diagrams/model_archi.drawio.pdf}
    \label{fig:model:archi}
  }
  \subfigure[Energy Prediction.]{
    \includegraphics[width=0.26125\linewidth]{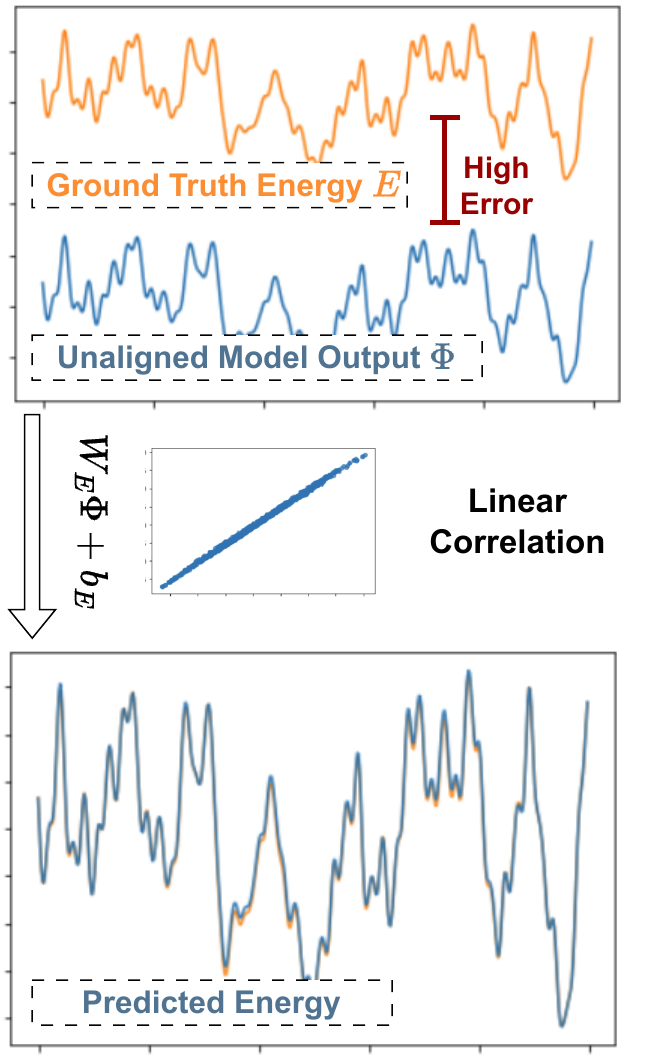}
    \label{fig:model:energy_prediction}
  }
  \caption{Illustration of the model architecture. Figure~\ref{fig:model:archi}: Equivariant Transformer learns to output \emph{pseudo-potential energy}, whose negative gradient is optimized to be atomic forces. The atomic forces can be used for MD simulations with an integrator. Figure \ref{fig:model:energy_prediction}: The pseudo-potential energy is not directly optimized to predict the ground truth but can approximate the latter with a linear transform. }
  \label{fig:model}
\end{figure*}
}

%% file: data.tex
\newcommand{\molwidth}{0.04\textwidth}

\begin{table}[h!]
  \input{tables/polymer_table}
  \caption{Polymers in our used \ourdata~dataset for training \ours.  We follow \cite{huan2017universal} and classify polymers into 4 broad categories: cycloalkanes, lactones, ethers, and others.  }
  \label{table:data}
\end{table}
\section{Dataset}
\label{sec:data}


Our PolyGET model is trained on a DFT-generated dataset  \cite{tran2022toward} for  computing ring-opening enthalpy ($\Delta H^{\rm ROP}$), which is a crucial thermodynamic concept  governing ring-opening polymerization. This process involves breaking a ring of cyclic monomers and adding the ``opened'' monomer to a long chain, which forms a polymer chain. The data were generated by utilizing MD simulations of multiple monomer and polymer models at a uniform level of DFT \cite{DFT1, DFT2} computations. More details about our data generation process can be found in Refs. \cite{tran2022toward, huan2016polymer}.

Given a cyclic monomer, several polymers models were created using Polymer Structure Predictor (PSP) package \cite{sahu2022polymer}. Each of them was obtained by multiplying the monomer with a small integer $L$, e.g., $L = 3, 4, 5$, and $6$, forming a loop of size $L$ (larger loops are better models of polymers). For each monomer or polymer models, about 10 or more maximally diversified configurations were selected as the initial configurations of the DFT-based MD simulations.

As all of our models are non-periodic, we used the $\Gamma$-point version of Vienna \textit{Ab initio} Simulation Package ({\sc vasp}),\cite{vasp3,vasp4} employing a basis set of plane waves with kinetic energy up to 400 eV to represent the Kohn-Sham orbitals. The ion-electron interactions were computed using the projector augmented wave (PAW) method \cite{PAW} while the exchange-correlation (XC) energies were computed using the generalized gradient approximation Perdew-Burke-Ernzerhof (PBE) functional.\cite{PBE}

We show the polymers used in this study in Table~\ref{table:data}, which displays all 24 polymer types which have been broadly classified into cycloalkanes, ethers, lactones, and others, following the classification used in \cite{tran2022toward}. Averagely, the dataset includes 10 DFT trajectories for both monomers and polymers. The polymers are formed by polymerizing the monomers, resulting in a comprehensive collection of data for each polymer type.
The average DFT trajectory lengths for polymers up to 6-loop are listed in Table~\ref{tab:avg_length}. In total, we have 1311 DFT trajectories and 6,552,624 molecular conformations.

\begin{table}[h!]
  \centering
  \begin{tabular}{c|cccccc}
    &  Monomer & 2-loop & 3-loop & 4-loop & 5-loop & 6-loop \\ \hline
    \# Trajectory. & 233 & 20 & 280 & 270 & 282 & 220 \\
    Avg. Length & 9526.31 & 8464.15 & 6182.98 & 4293.98 & 3180.02 & 1614.87
  \end{tabular}
  \caption{Average trajectory length w.r.t. polymer size. }
  \label{tab:avg_length}
\end{table}

%% file: tables/polymer_table.tex
\begin{tabular}{lllll}
\hline
Index         & Type                         & \# Atoms & Monomer                                                            & Polymer                                                            \\ \hline
\textbf{CK1}  & cyclopropane                 & 9       & \includegraphics[width=\molwidth]{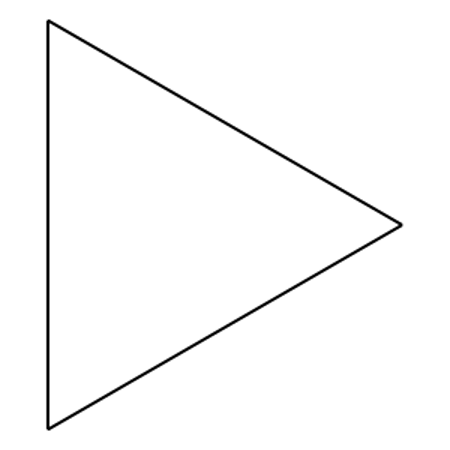}   & \includegraphics[width=\molwidth]{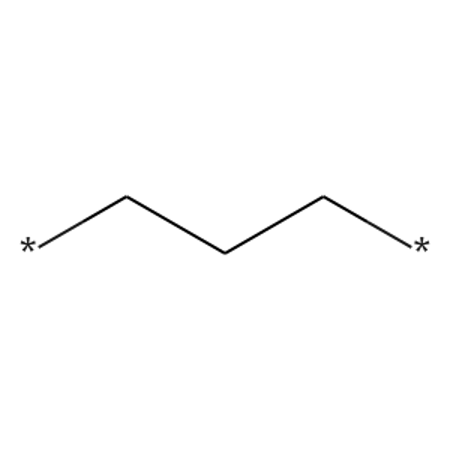}   \\
\textbf{CK2}  & cyclobutane                  & 12       & \includegraphics[width=\molwidth]{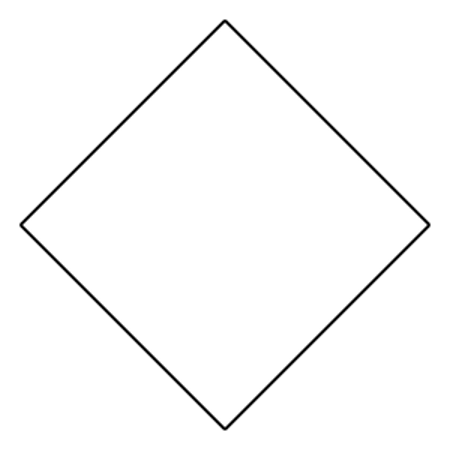}   & \includegraphics[width=\molwidth]{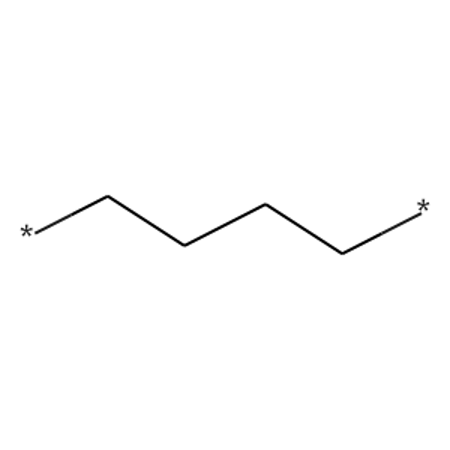}   \\
\textbf{CK3}  & cyclopentane                 & 15      & \includegraphics[width=\molwidth]{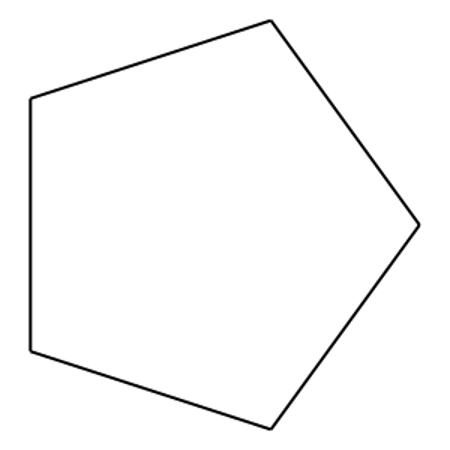}   & \includegraphics[width=\molwidth]{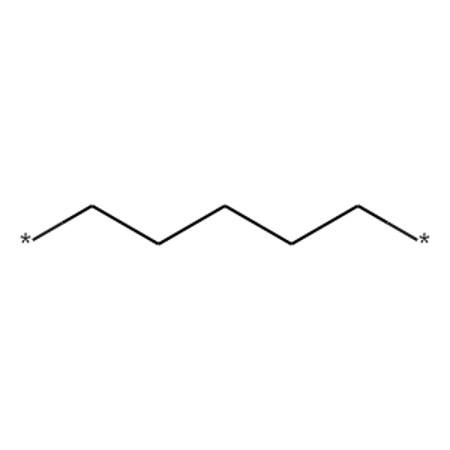}   \\
\textbf{CK4}  & cyclohexane                  & 18       & \includegraphics[width=\molwidth]{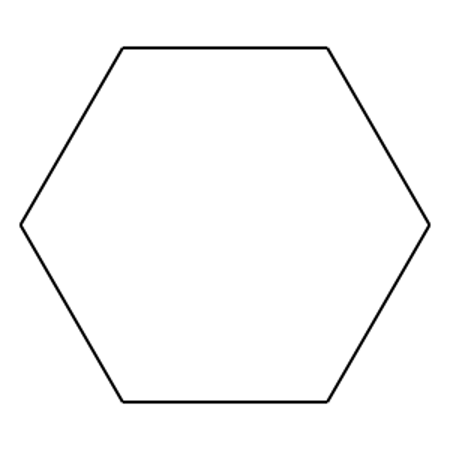}  & \includegraphics[width=\molwidth]{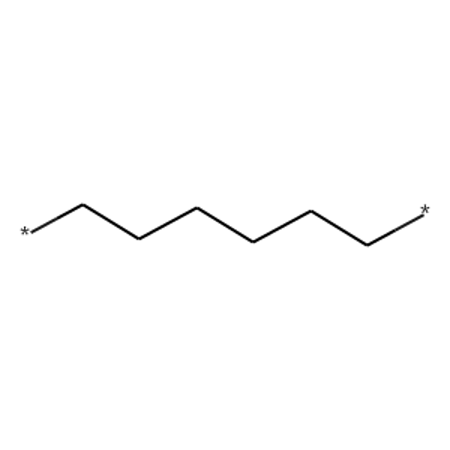}  \\
\textbf{CK5}  & cycloheptane                 & 21       & \includegraphics[width=\molwidth]{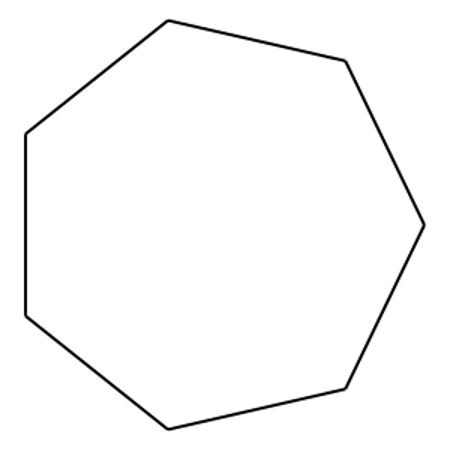}  & \includegraphics[width=\molwidth]{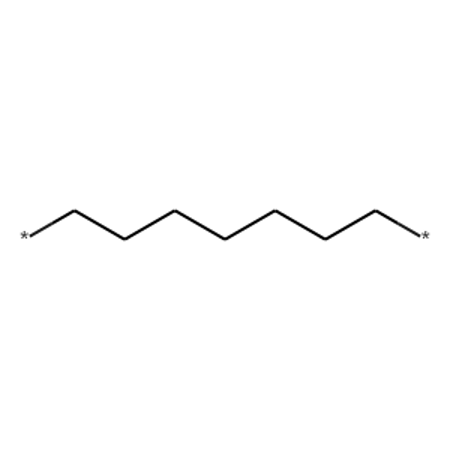}  \\
\textbf{CK6}  & cycloctane                   & 24      & \includegraphics[width=\molwidth]{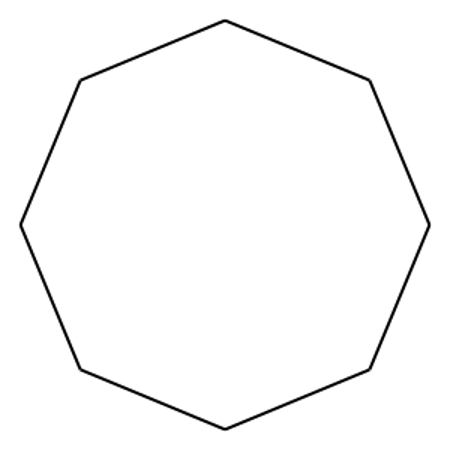}  & \includegraphics[width=\molwidth]{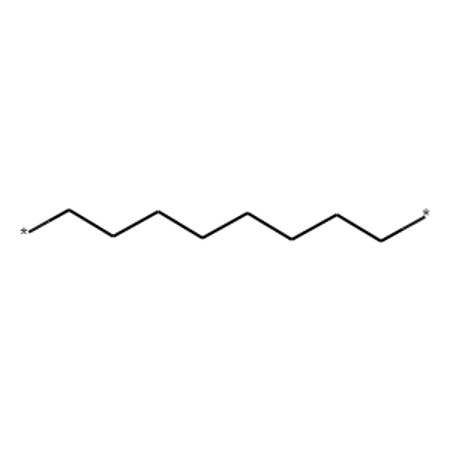}  \\ \hline
\textbf{ETH}  & ethylene oxide               & 7       & \includegraphics[width=\molwidth]{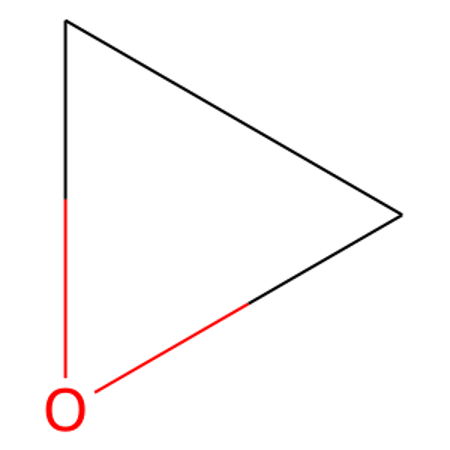}   & \includegraphics[width=\molwidth]{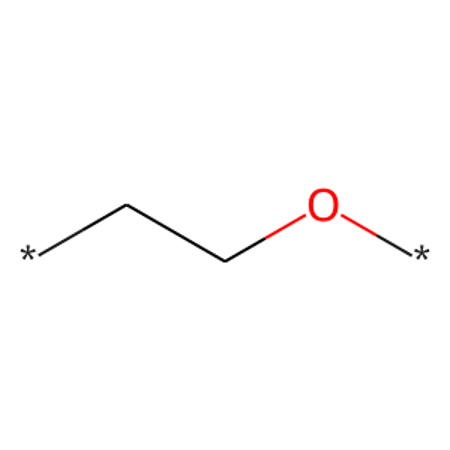}   \\ \hline
\textbf{LAC1} & \(\gamma\)-butyrolactone              & 9       & \includegraphics[width=\molwidth]{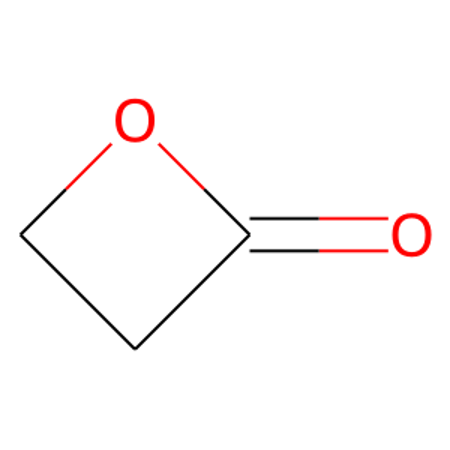}   & \includegraphics[width=\molwidth]{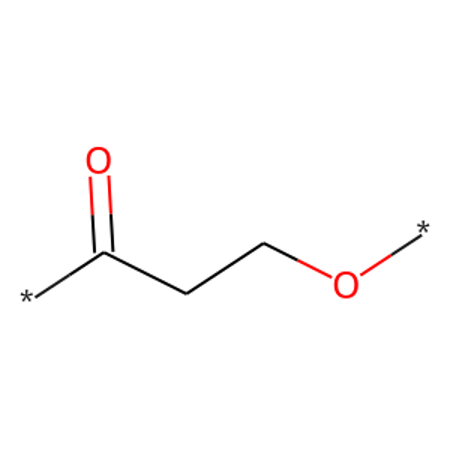}   \\
\textbf{LAC2} & \(\gamma\)-butyrolactone              & 12       & \includegraphics[width=\molwidth]{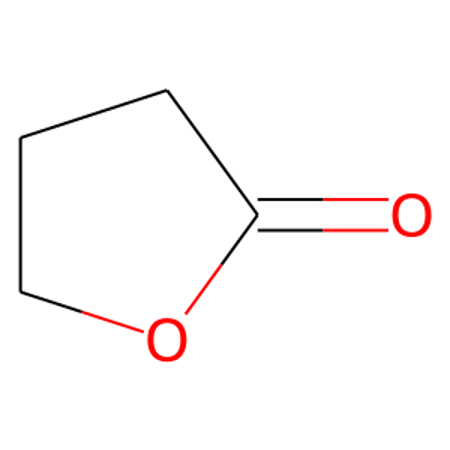}    & \includegraphics[width=\molwidth]{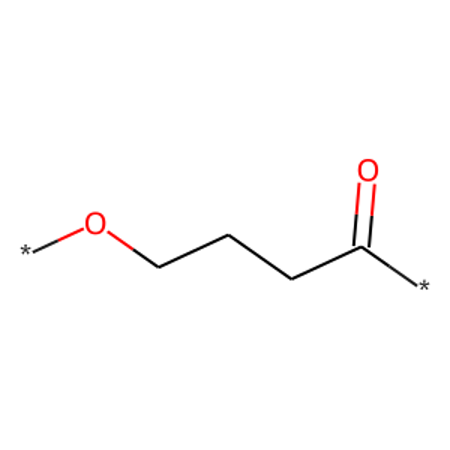}    \\
\textbf{LAC3} & 1,4-dioxan-2-one             & 13      & \includegraphics[width=\molwidth]{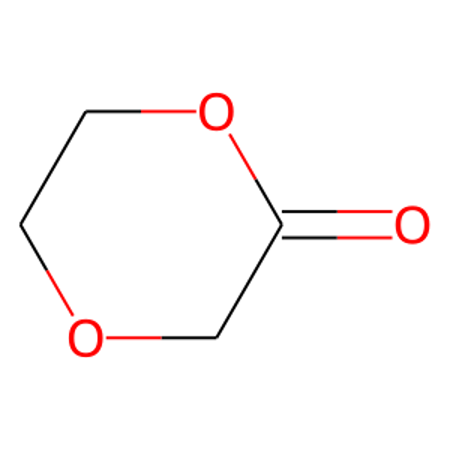}   & \includegraphics[width=\molwidth]{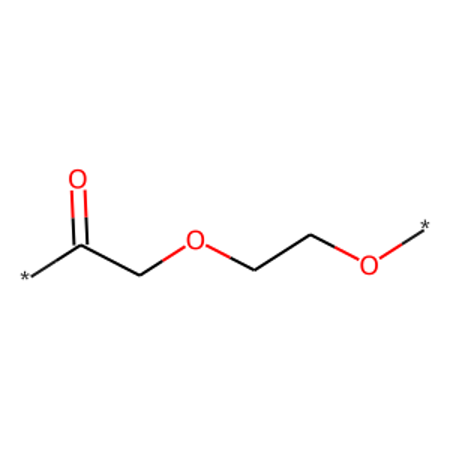}   \\
\textbf{LAC4} & \(\delta\)-valerolactone              & 15      & \includegraphics[width=\molwidth]{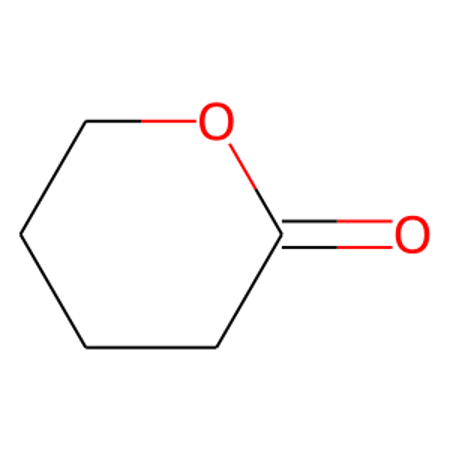}   & \includegraphics[width=\molwidth]{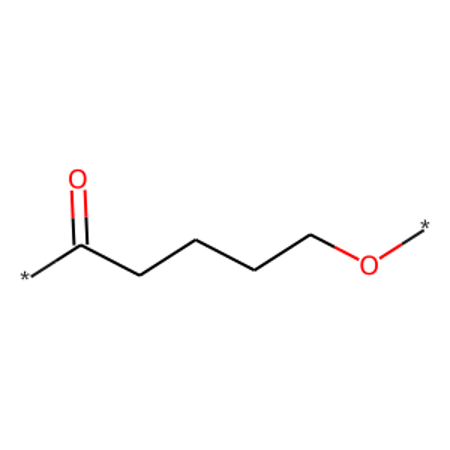}   \\
\textbf{LAC5} & 3-methyl-1,4-dioxan-2-one    & 16      & \includegraphics[width=\molwidth]{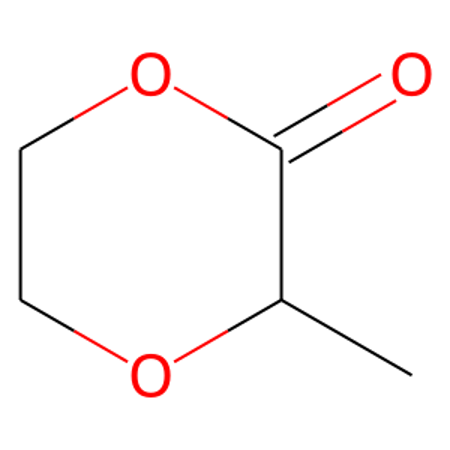}    & \includegraphics[width=\molwidth]{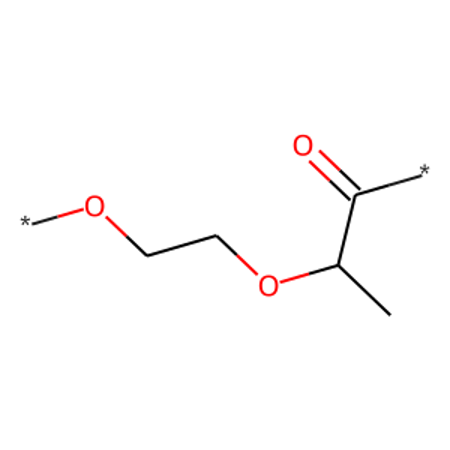}    \\
\textbf{LAC6} & \(\beta\)-methyl-\(\delta\)-valerolactone     & 18      & \includegraphics[width=\molwidth]{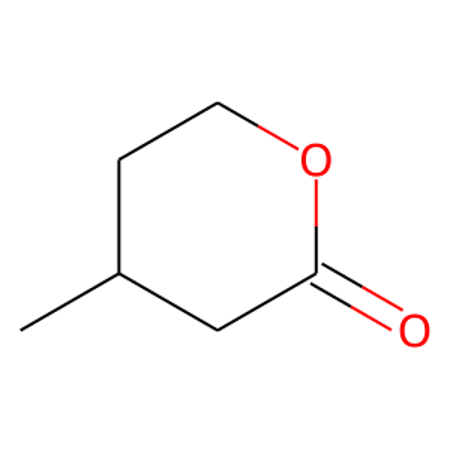}    & \includegraphics[width=\molwidth]{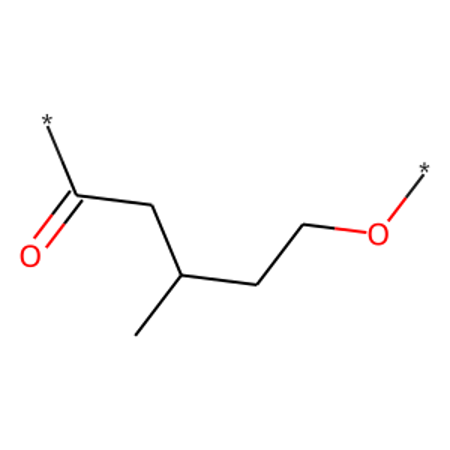}    \\
\textbf{LAC7} & \(\delta\)-caprolactone               & 18      & \includegraphics[width=\molwidth]{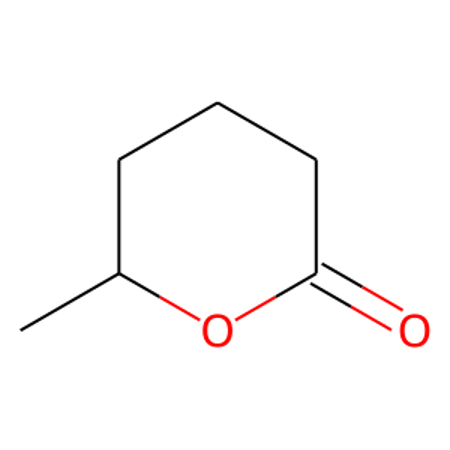}    & \includegraphics[width=\molwidth]{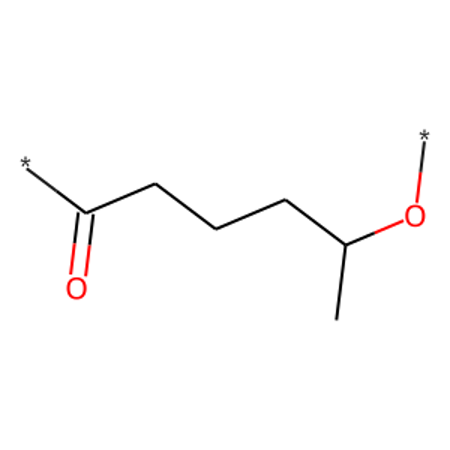}    \\
\textbf{LAC8} & \(\delta\)-decalactone                & 30      & \includegraphics[width=\molwidth]{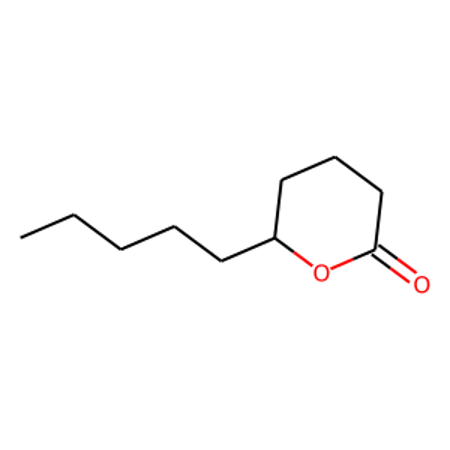}   & \includegraphics[width=\molwidth]{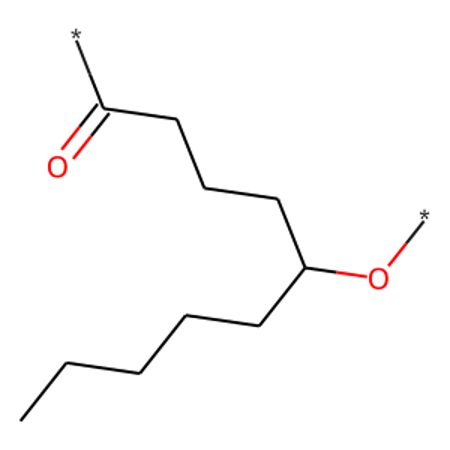}   \\
\textbf{LAC9} & (-)-Menthide                 & 30      & \includegraphics[width=\molwidth]{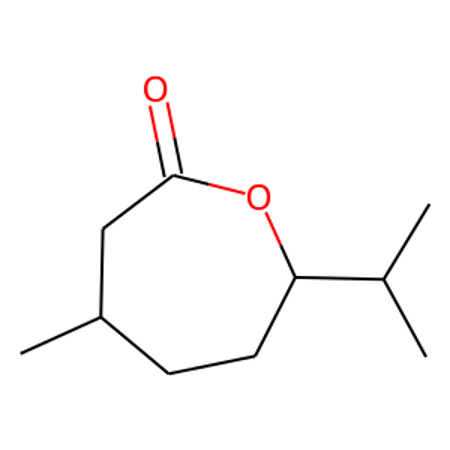}    & \includegraphics[width=\molwidth]{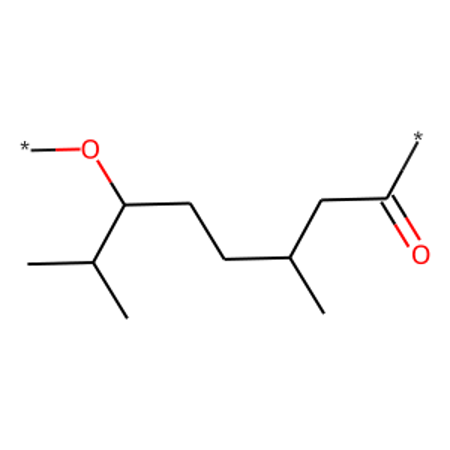}    \\ \hline
\textbf{OTH1} & n-alkane sub \(\delta\)-valerolactone & 18      & \includegraphics[width=\molwidth]{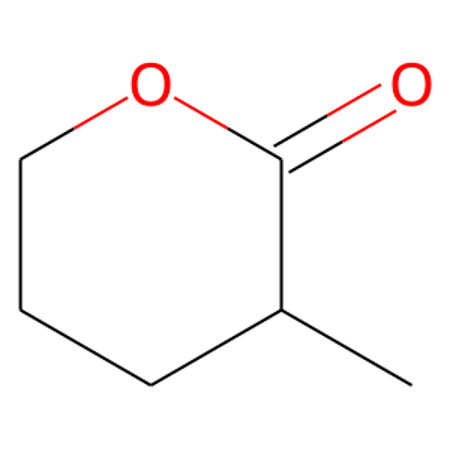} & \includegraphics[width=\molwidth]{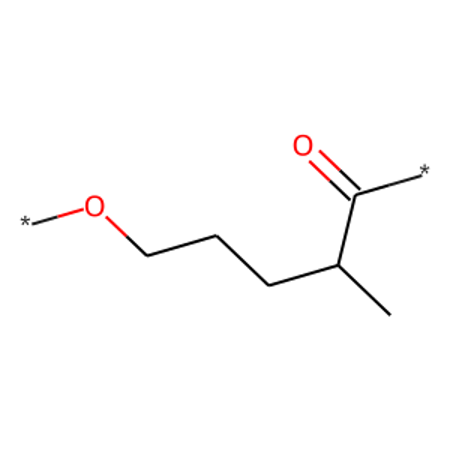} \\
\textbf{OTH2} & \(\alpha\)-Methylene-\(\gamma\)-butyrolactone  & 13      & \includegraphics[width=\molwidth]{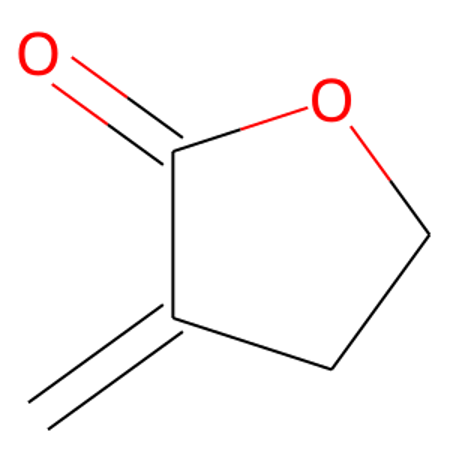} & \includegraphics[width=\molwidth]{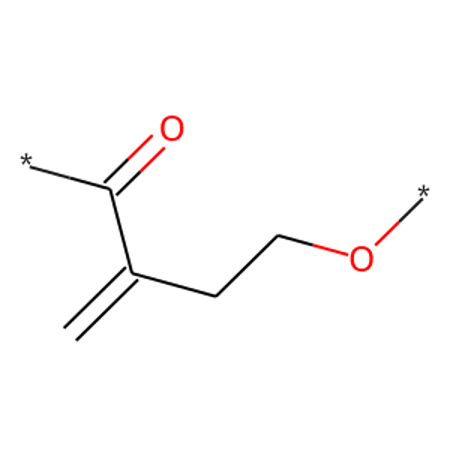} \\
\textbf{OTH3} & n-alkane sub \(\delta\)-valerolactone & 21      & \includegraphics[width=\molwidth]{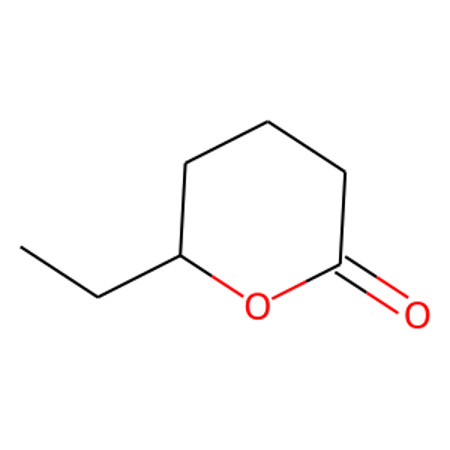} & \includegraphics[width=\molwidth]{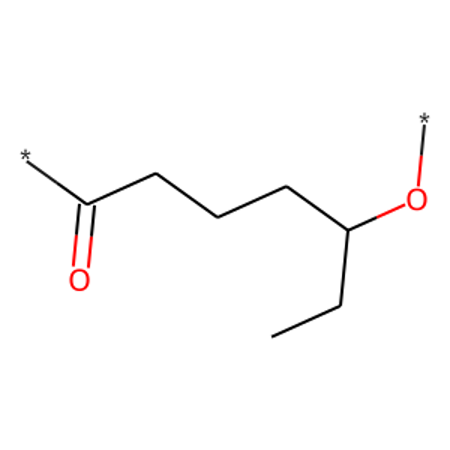} \\
\textbf{OTH4} & n-alkane sub \(\delta\)-valerolactone & 24      & \includegraphics[width=\molwidth]{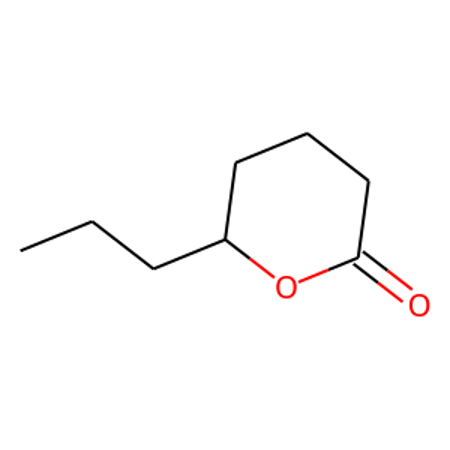} & \includegraphics[width=\molwidth]{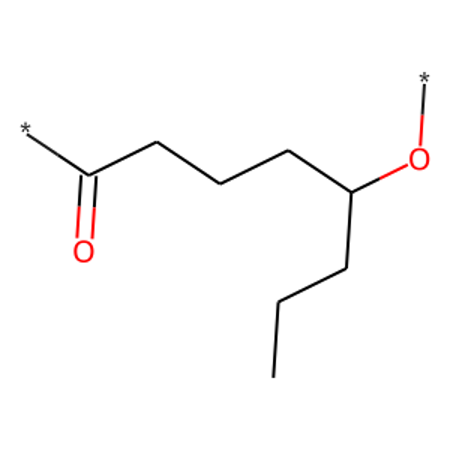} \\
\textbf{OTH5} & n-butyl \(\delta\)-valerolactone      & 27      & \includegraphics[width=\molwidth]{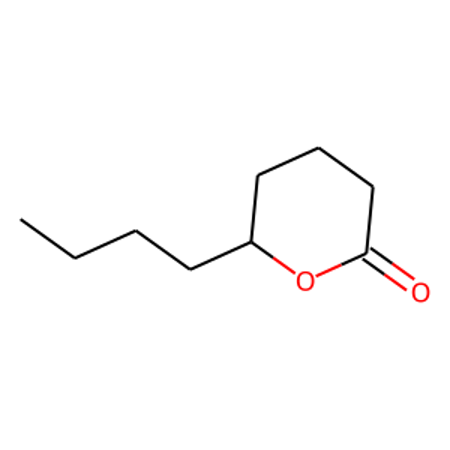} & \includegraphics[width=\molwidth]{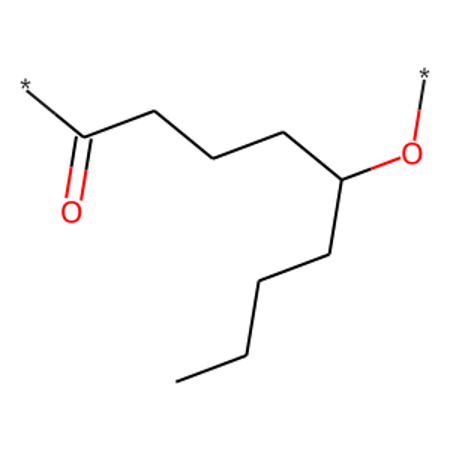} \\
\textbf{OTH6} & n-alkane sub \(\delta\)-valerolactone & 30      & \includegraphics[width=\molwidth]{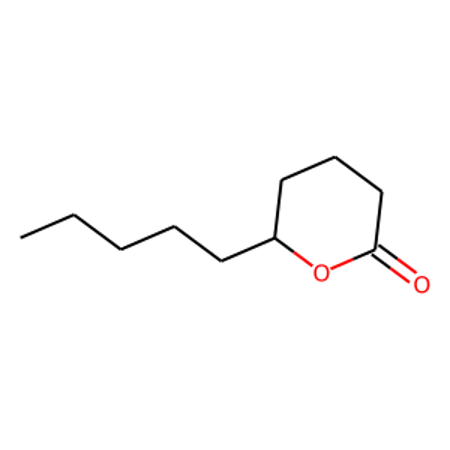} & \includegraphics[width=\molwidth]{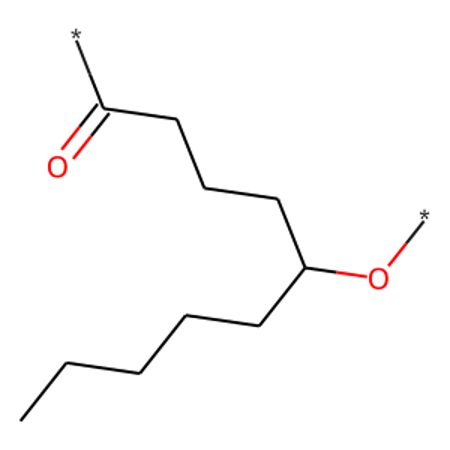} \\
\textbf{OTH7} & n-alkane sub \(\delta\)-valerolactone & 33      & \includegraphics[width=\molwidth]{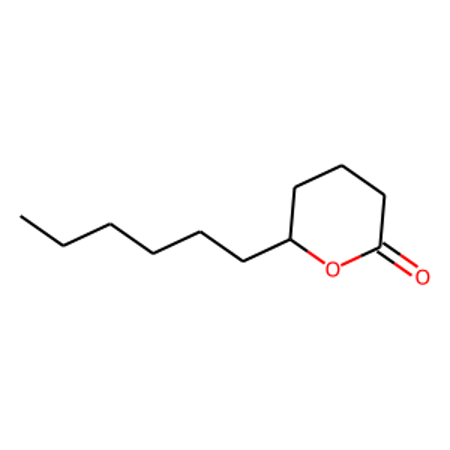} & \includegraphics[width=\molwidth]{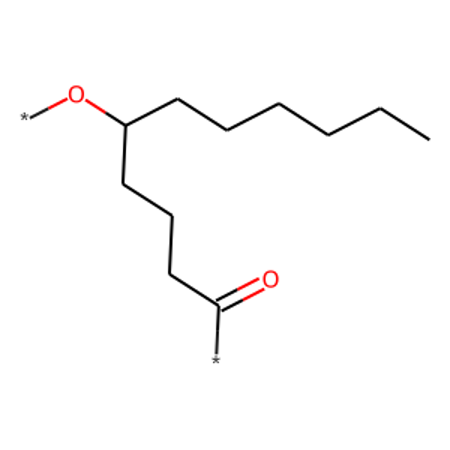} \\
\textbf{OTH8} & n-alkane sub \(\delta\)-valerolactone & 42      & \includegraphics[width=\molwidth]{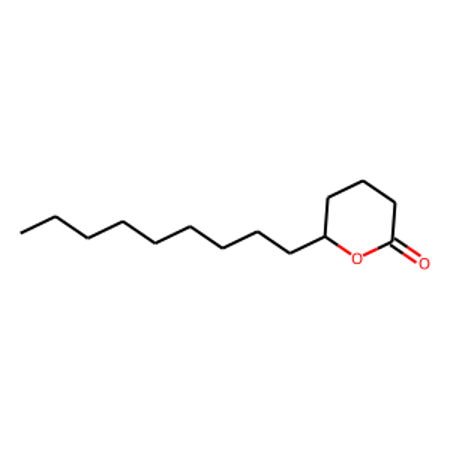} & \includegraphics[width=\molwidth]{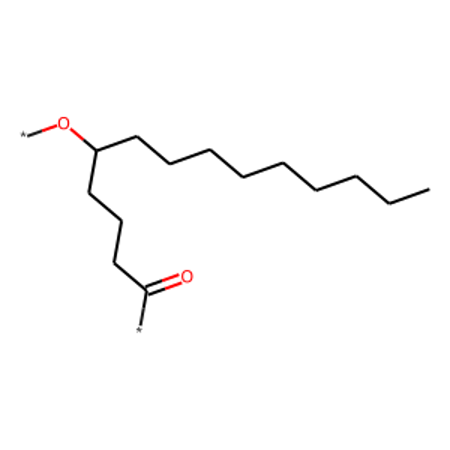} \\ \hline
\end{tabular}

%% file: short_versions/experiment.tex
\section{Results}
\label{sect:result}
\newcommand{\widthere}{0.22}
\begin{figure}[h!]
  \centering
  \subfigure[Forces Correlation for 15 Loop CK6.]{
    \includegraphics[width=\widthere\linewidth]{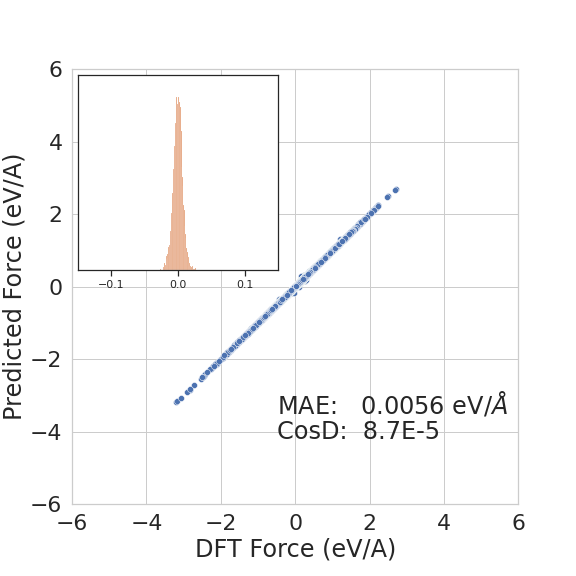}
  }
  \subfigure[Forces Correlation for 15 Loop OTH4.]{
    \includegraphics[width=\widthere\linewidth]{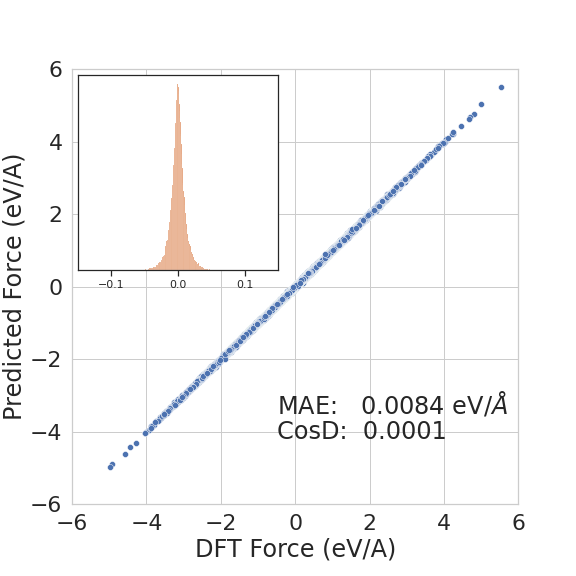}
  }
  \subfigure[Forces Correlation for 15 Loop LAC2.]{
    \includegraphics[width=\widthere\linewidth]{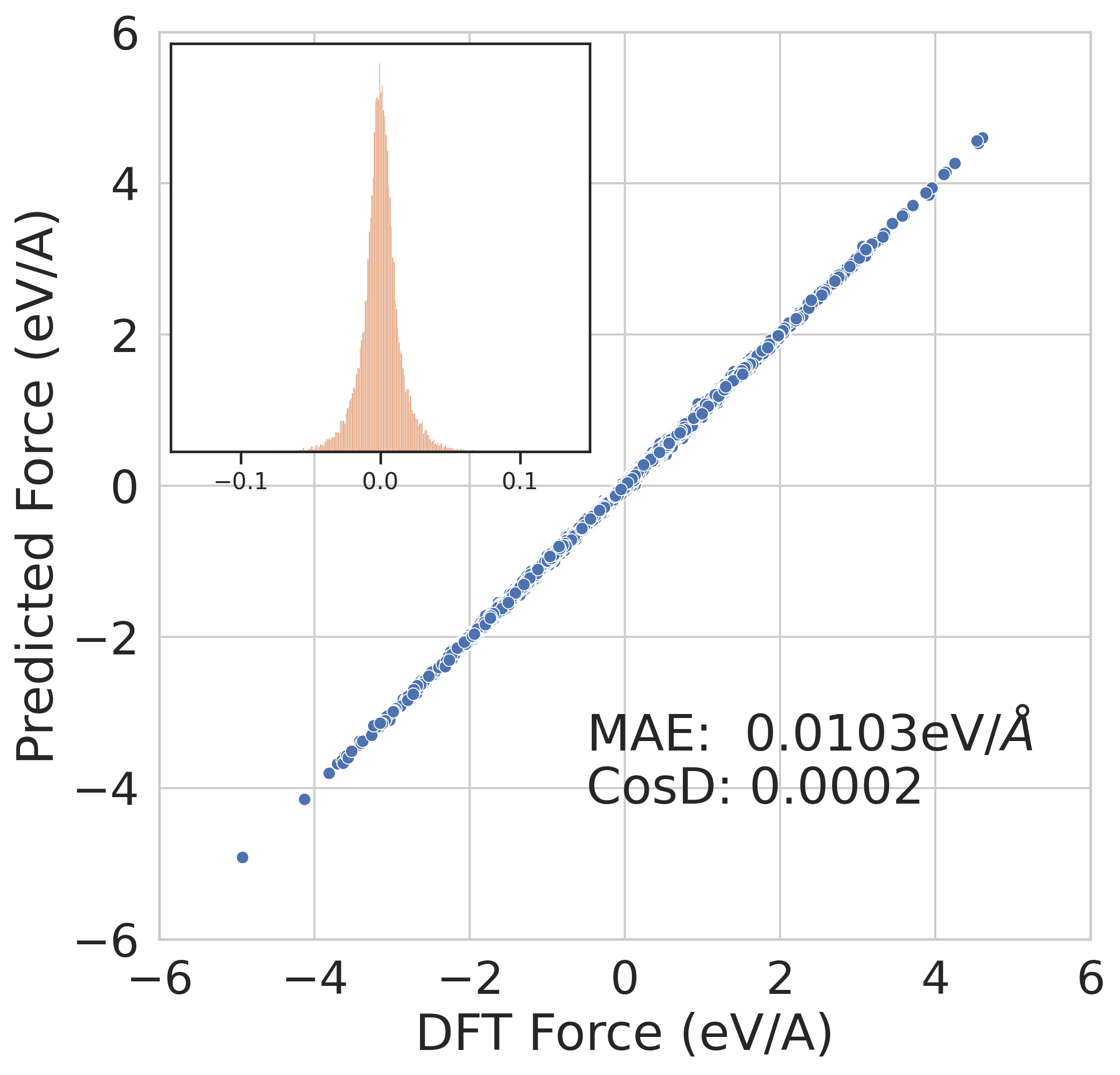}
  }
  \subfigure[Forces Correlation for 15 Loop LAC5.]{
    \includegraphics[width=\widthere\linewidth]{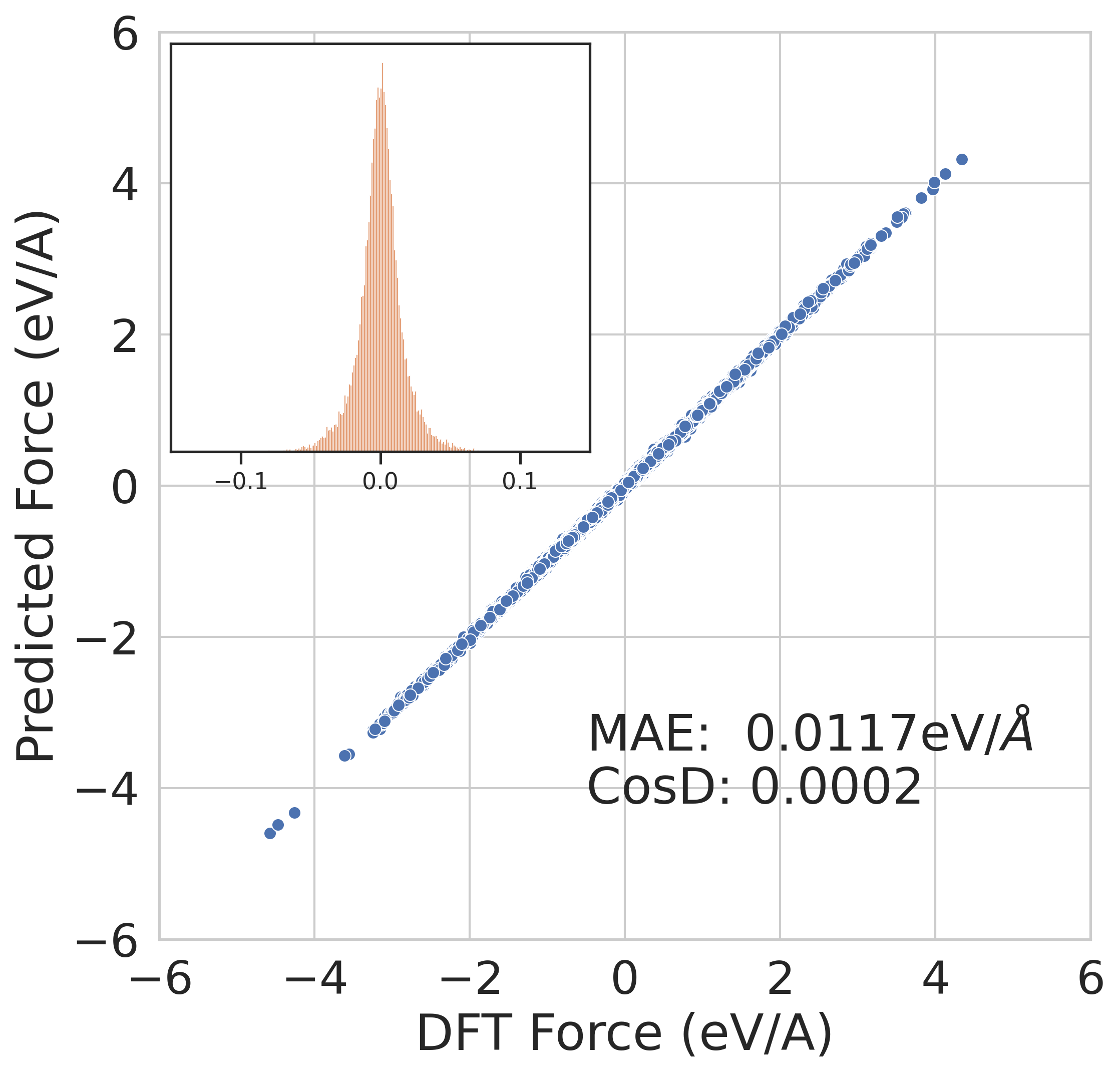}
  }

  \subfigure[RMSD and Potential Energy for 15 Loop CK6.]{
    \includegraphics[width=\widthere\linewidth]{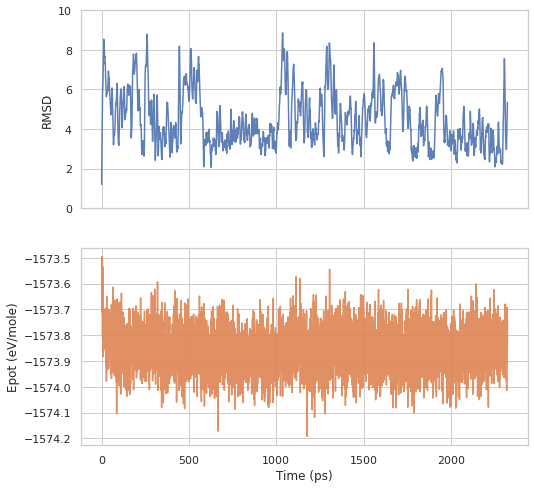}
  }
  \subfigure[RMSD and Potential Energy for 15 Loop OTH4.]{
    \includegraphics[width=\widthere\linewidth]{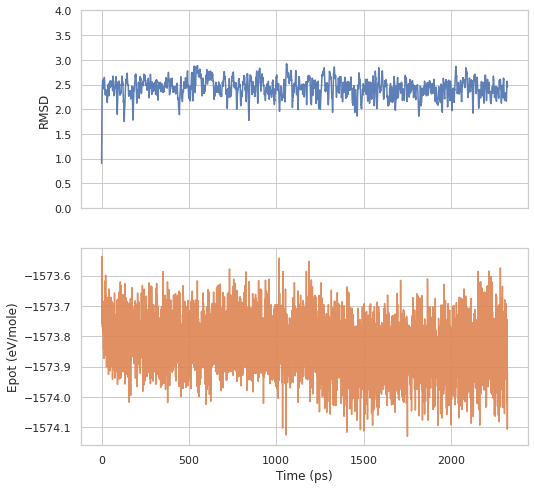}
  }
  \subfigure[RMSD and Potential Energy for 15 Loop LAC2.]{
    \includegraphics[width=\widthere\linewidth]{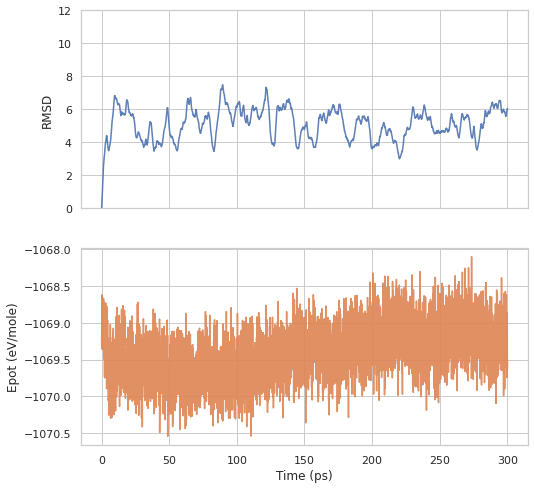}
  }
  \subfigure[RMSD and Potential Energy for 15 Loop LAC5.]{
    \includegraphics[width=\widthere\linewidth]{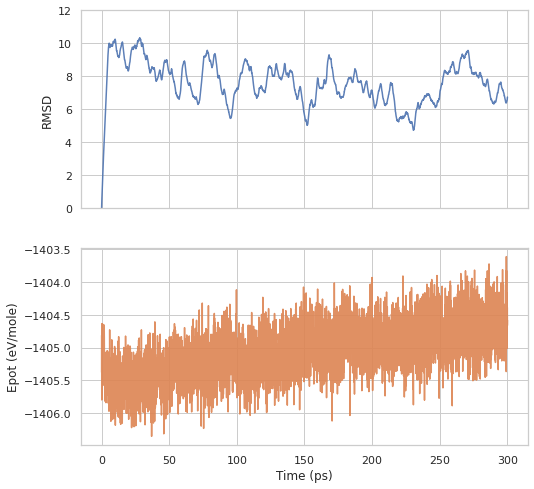}
  }

  \subfigure[Dynamics Illustration for CK6.]{
    \includegraphics[width=\widthere\linewidth]{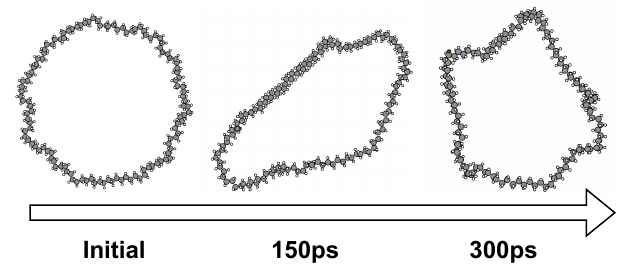}
  }
  \subfigure[Dynamics Illustration for OTH4.]{
    \includegraphics[width=\widthere\linewidth]{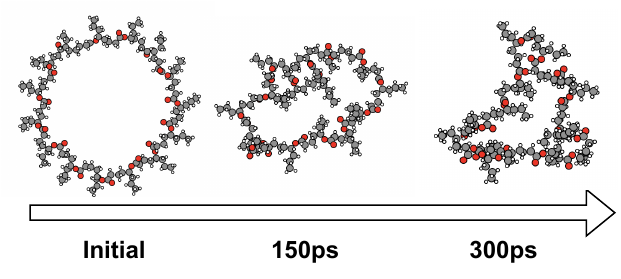}
  }
  \subfigure[Dynamics Illustration for LAC2.]{
    \includegraphics[width=\widthere\linewidth]{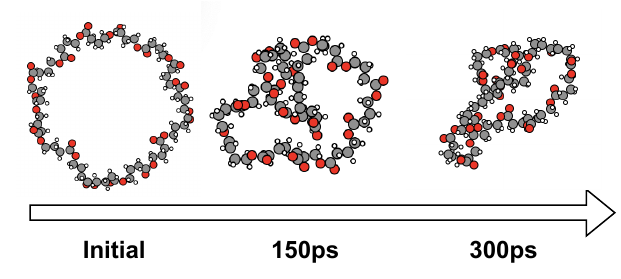}
  }
  \subfigure[Dynamics Illustration for LAC5.]{
    \includegraphics[width=\widthere\linewidth]{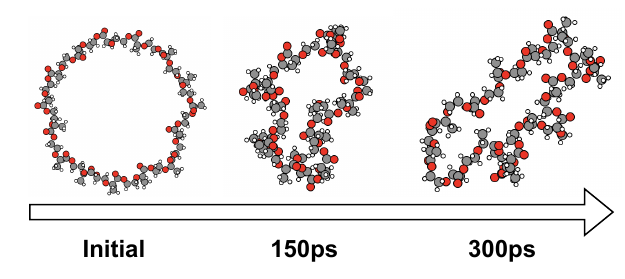}
  }
  \caption{Accuracy and Robustness of \ours~ in MD simulations for large polymers. We select 15 loop CK6 (cycloctane), OTH4 (n-alkane substituted \(\delta\)-valerolactone), LAC2 (\(\gamma\)-butyrolactone) and LAC5 (3-methyl-1,4-dioxan-2-one) as examples. In (a)-(d) it is observed that \ours~forces are highly correlated with DFT forces during its own MD simulations. (e)-(h) shows the RMSD and potential energy curves of \ours~ during MD simulations. The curves suggest that \ours~simulations experience equilibration. (i)-(j) presents a visualization of sample conformations captured at various time points during the simulation process, which shows \ours could effectively explore and reveal the dynamic changes in polymer structures over time.}
  \label{fig:large_polymer}
\end{figure}

In this section, we present the empirical performance of \ours~on \ourdata.
First, we report on the simulation performance of \ours~on large 15-loop
unseen polymers in Section \ref{sec:exp:sim}, demonstrating the accuracy,
robustness, and speed of \ours ~for large polymer simulation. Then, we
explore the multi-molecule training performance of \ours~in Section
\ref{sec:exp:multimolecule}. We show that our training paradigm is crucial
in enhancing force prediction accuracy and the generalization capabilities
of the model.

\input{exp_sim.tex}

\begin{figure*}[h!]
  \subfigure[SchNet]{\includegraphics[width=.24\linewidth]{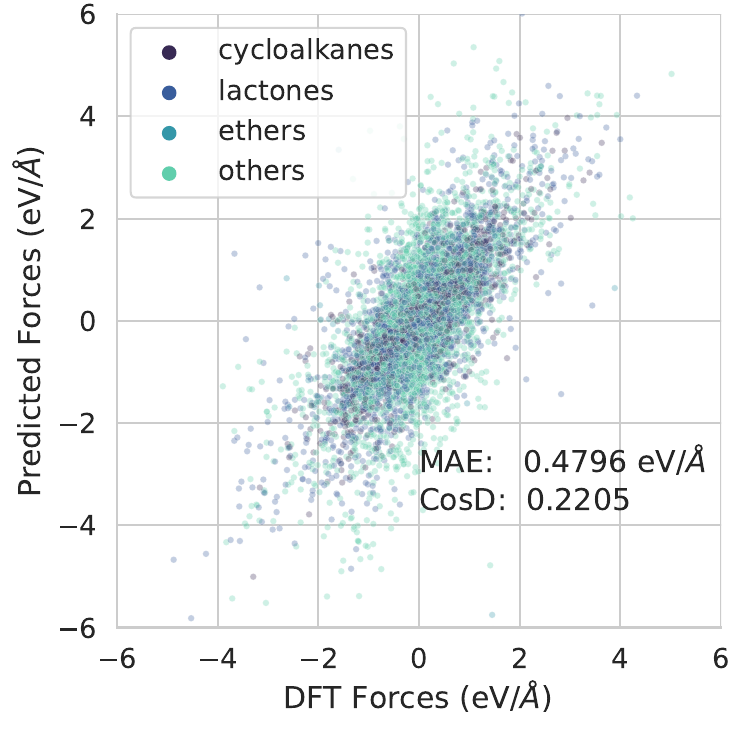}}
  \subfigure[NequIP]{\includegraphics[width=.24\linewidth]{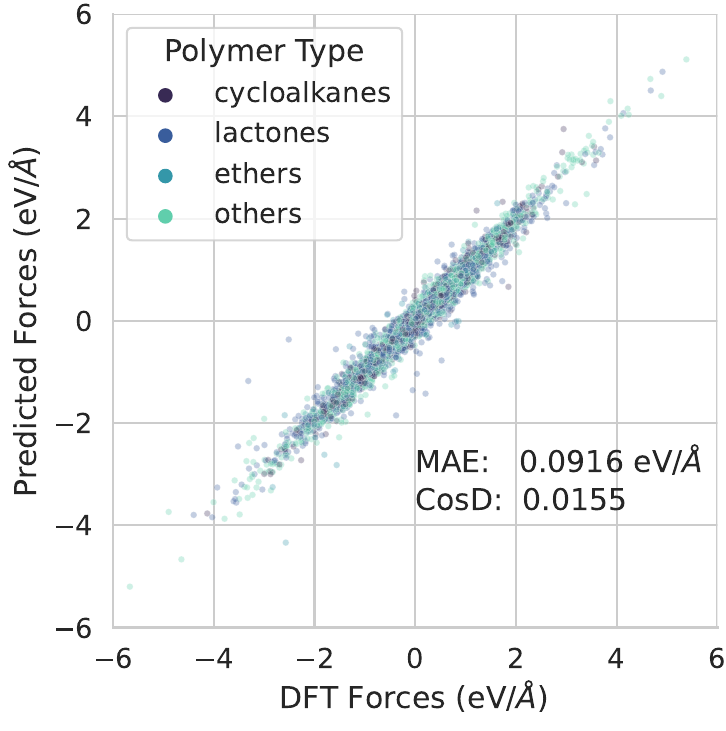}}
  \subfigure[TorchMDNet]{\includegraphics[width=.24\linewidth]{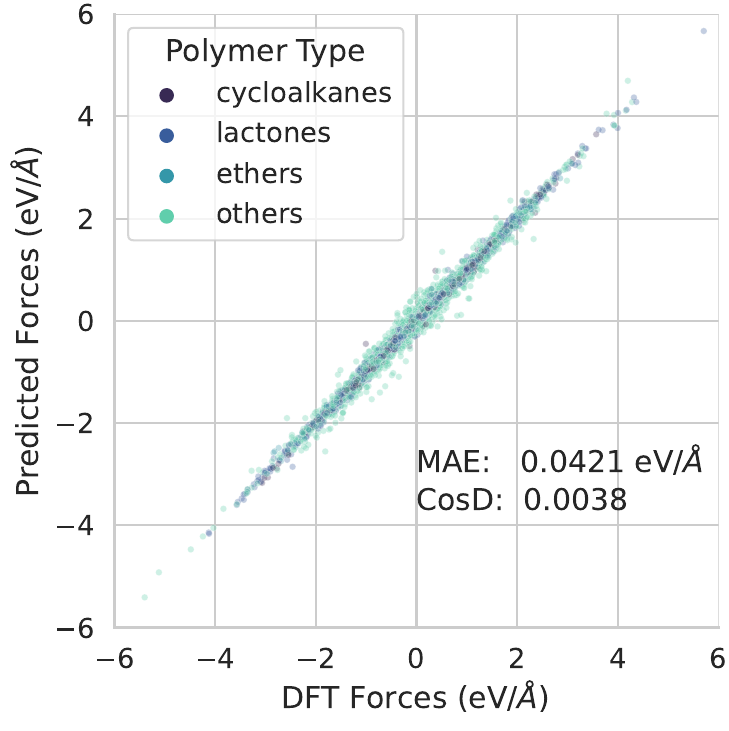}}
  \subfigure[\ours]{\includegraphics[width=.24\linewidth]{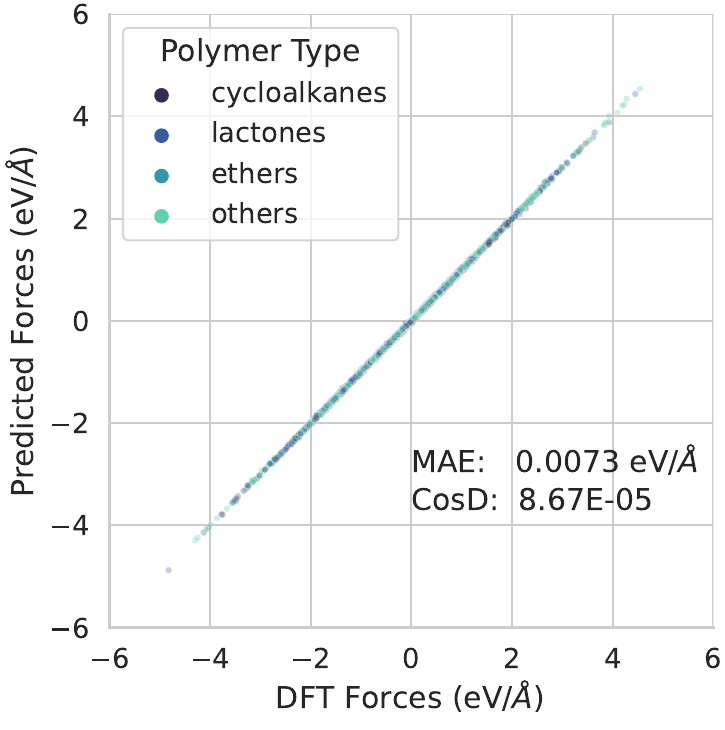}}
  \caption{Performance of deep learning forcefield, including SchNet, NequIP, TorchMDNet, and ~\ours~on multi-molecule training. Two metrics, Mean Absolute Error (MAE) and Cosine Distance (CosD) are reported. On the left, (a) shows the MAE comparison between models across 4 types of monomers: cycloalkanes, lactones, ethers, and others, as in Table~\ref{table:data}. On the right, (b) - (e) shows the scatter plots of DFT forces compared with predicted forces for the 4 methods. }
  \label{fig:multimodel}
\end{figure*}

\subsection{Effectiveness of Multi-molecule Training}
\label{sec:exp:multimolecule}
In this subsection, we show that the multi-molecule training paradigm can crucially improve the prediction performance of \ours. To this end, we compare \ours~ with 3 state-of-the-art machine learning forcefields models: SchNet~\cite{schutt2018schnet}, NequIP~\cite{batzner20223}, and TorchMDNet~\cite{tholke2022torchmd}.
These methods feature different implementations of EGNN and have achieved accurate forcefield prediction on static DFT trajectories.

\begin{figure}[h!]

  \centering
  \subfigure[Multi-molecule vs. Single-molecule.]{
    \includegraphics[width=0.46\linewidth]{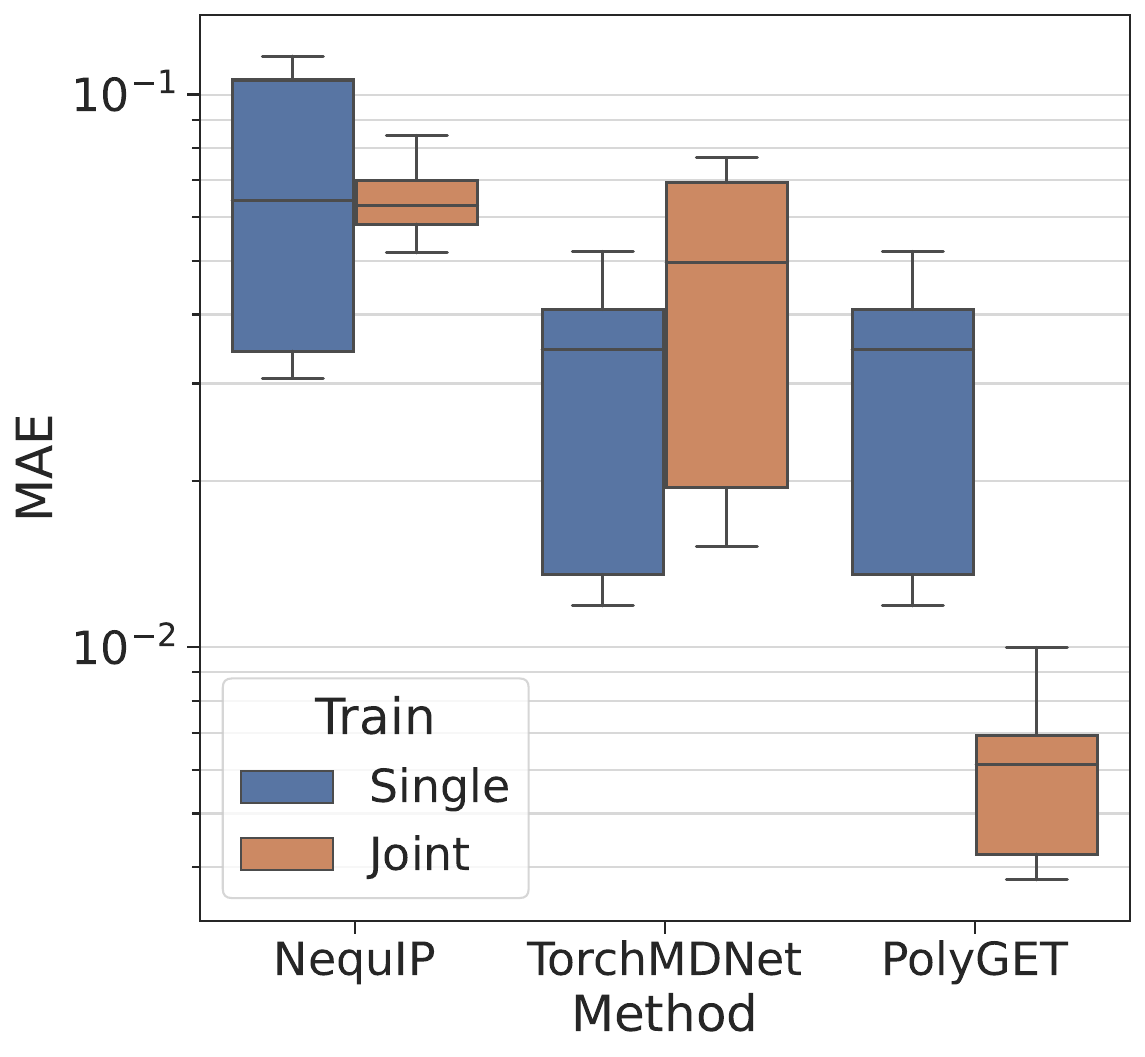}
    \label{fig:static:single_multi}
  }
  \subfigure[Extrapolation to Unseen Polymers.]{
    \includegraphics[width=0.46\linewidth]{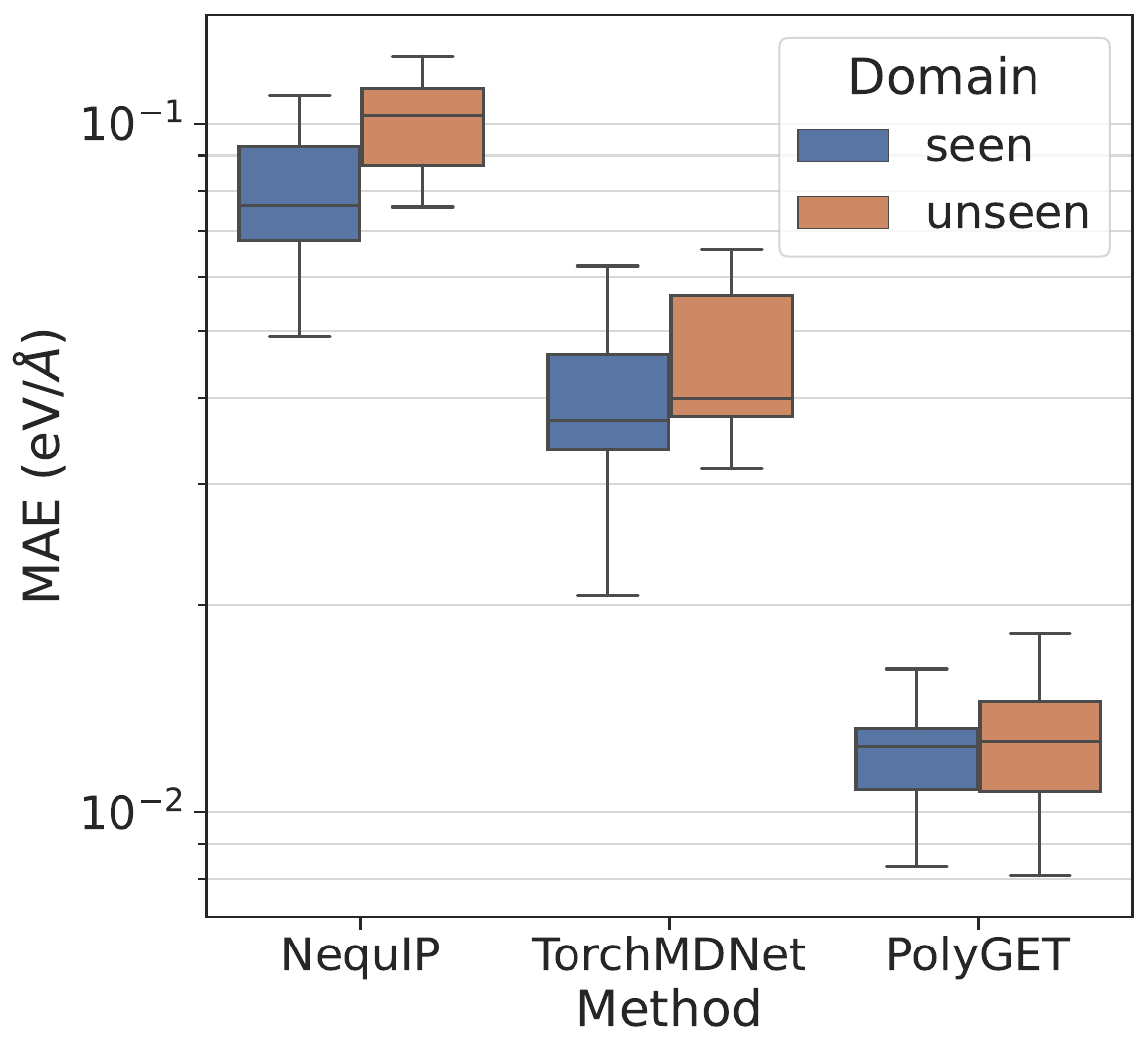}
    \label{fig:static:seen_unseen}
  }
  \centering
  \subfigure[Validation Forces MAE.]{
    \includegraphics[width=0.91\linewidth]{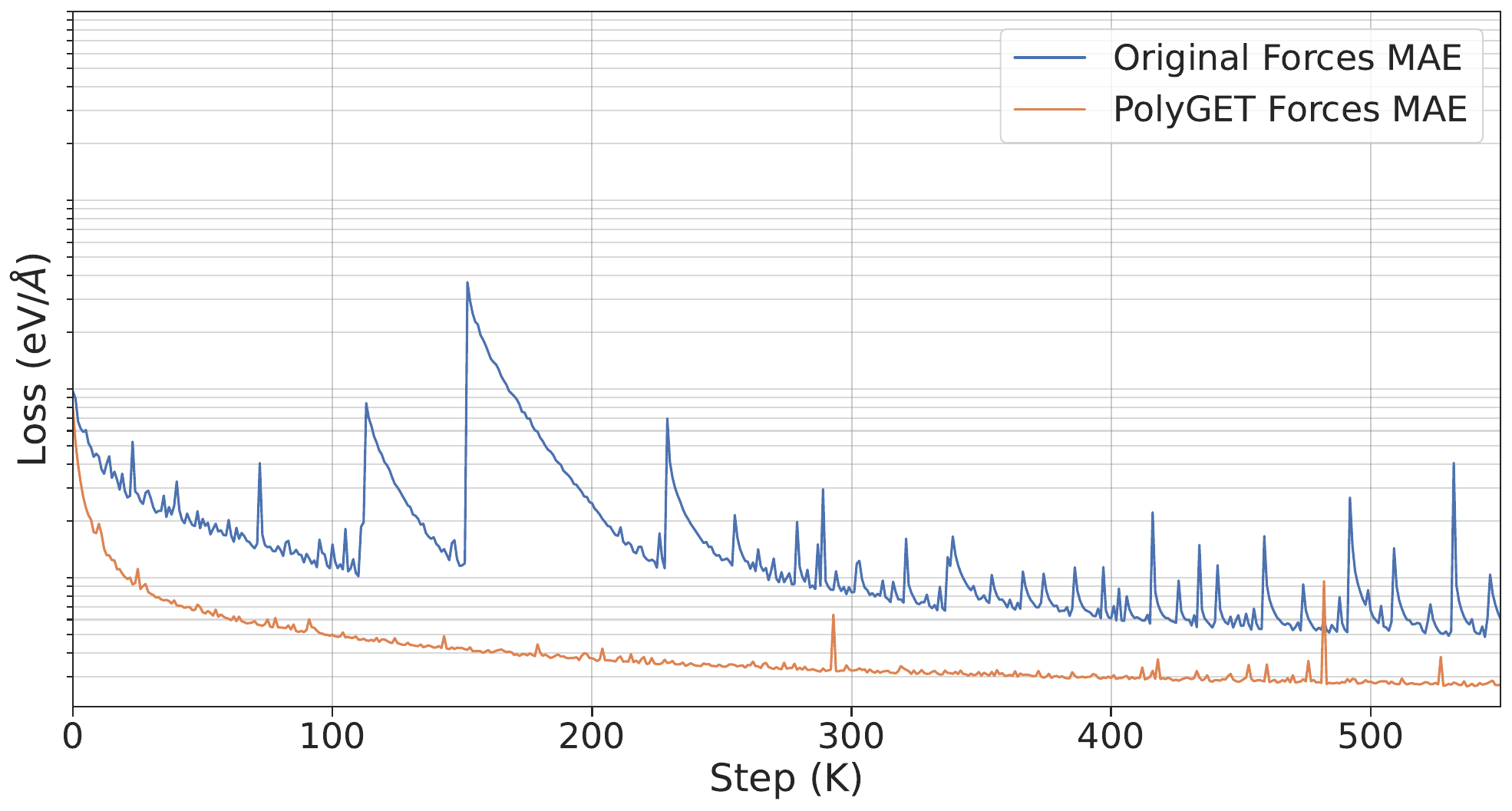}
    \label{fig:static:training}
  }
  \caption{Figure~\ref{fig:static:training}: validation forces MAE of the Equivariant Transformers during training, where ``Original'' refers to the original paradigm of joint energy and forces optimization, and ``PolyGET'' refers to our force-centered multi-molecule training paradigm. Figure~\ref{fig:static:single_multi}: comparison of forces accuracy of models trained on single molecules and multiple molecules. Figure~\ref{fig:static:seen_unseen}: we analyze \ours's extrapolation performance to polymers not seen during training.  }
  \label{fig:static_analysis}
\end{figure}

\paragraph*{Our model achieves state-of-the-art multi-molecule force prediction accuracy.}
To demonstrate this, we train \ours~ and the baseline models on all
polymers with loop sizes below 6 (Table~\ref{table:data}). We then evaluate
the force prediction performance of different models on unseen 6-loop
polymers of the same monomers. Table~\ref{fig:multimodel} reports the
results. As shown, our model \ours~attains 0.073 eV/\AA ~MAE and a near-zero
cosine distance compared to DFT forces, which is 6 times better than the
previous state-of-the-art model TorchMDNet.

\paragraph*{Our model benefits the most from diverse training data.} In
Figure~\ref{fig:static:single_multi}, we compared the performance of
baseline models when trained separately or jointly on different polymers.
For NequIP and TorchMDNet, joint training on all polymers brings no
significant improvement and could even harm force prediction accuracy. In
contrast, our model benefits from having multiple types of polymers in the
training data, significantly reducing force prediction errors. It is worth
noting that our model shares a similar Equivariant Transformer architecture
with TorchMDNet~\cite{tholke2022torchmd}. The performance gain is mainly
because of our force-centered training paradigm. As depicted in
Figure~\ref{fig:static:training}, This paradigm allows the validation MAE
on forces to decrease more consistently in multi-molecule training, leading
to a significantly improved model performance. Such results demonstrate the
effectiveness of our force-centric multi-molecule training paradigm.

\paragraph*{The multi-molecule paradigm improves model extrapolation.}
In Figure~\ref{fig:static:seen_unseen}, we report the baseline models' performance on ``unseen'' polymers compared to ``seen'' polymers. ``Seen'' polymers refer to polymers with loop sizes below 6 present in the training data, while ``unseen'' polymers pertain to 6-loop polymers. In line with previous multi-molecule performance observations, our model experiences minimal performance deterioration (0.012 eV/\AA ~to 0.013 eV/\AA) when applied to unseen polymers. In contrast, NequIP's force MAE drops from 0.080 eV/\AA ~to 0.103 eV/\AA, while TorchMDNet drops from 0.043 to 0.056. This result shows that the multi-molecule paradigm plays a vital role in enhancing the model's extrapolation capabilities.

\begin{figure}[h!]
    \centering
    \subfigure[Model Output vs DFT Potential Energy.]{
        \includegraphics[width=0.42\linewidth]{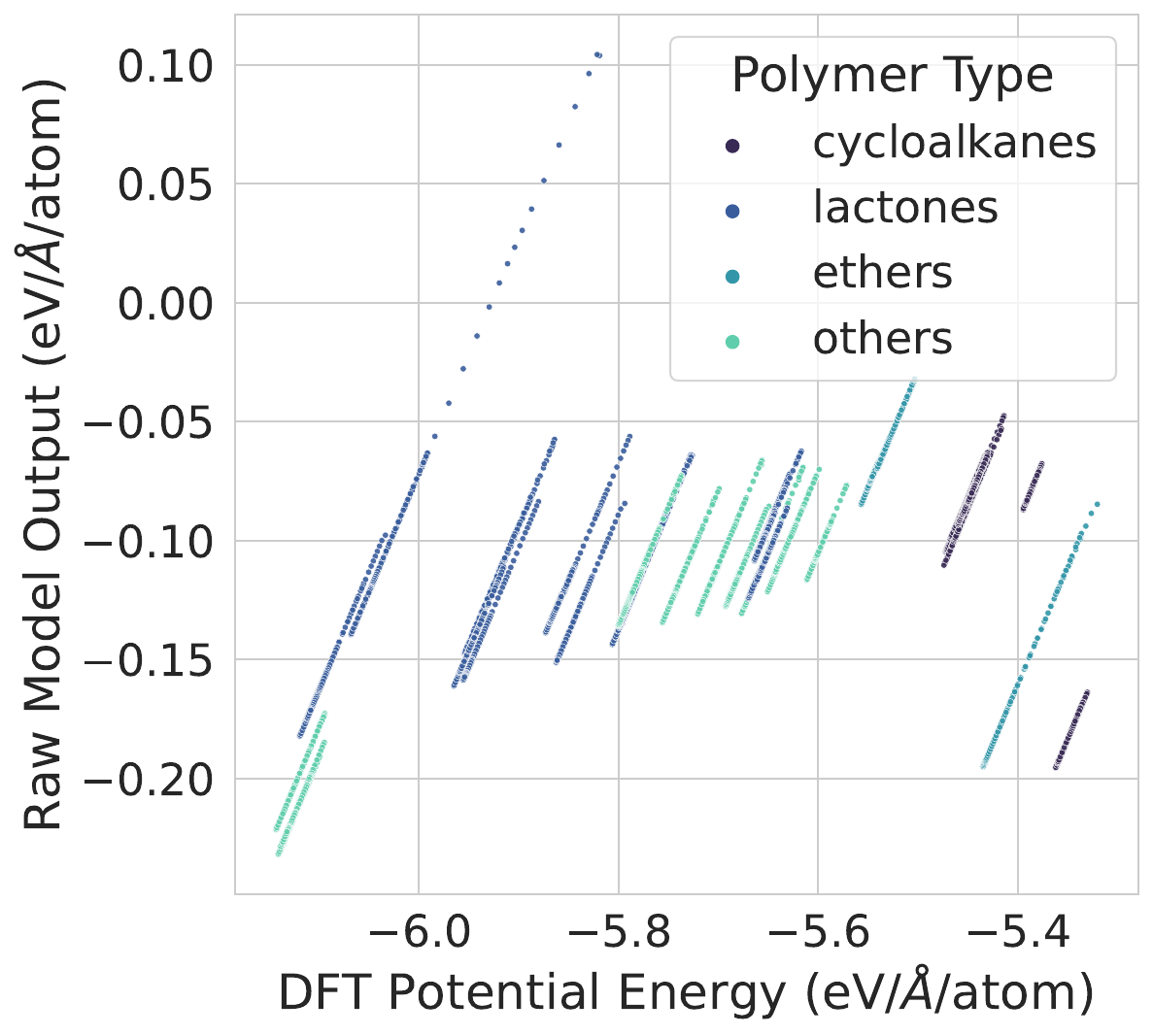}
        \label{fig:exp:unadjusted_energy}
    }
    \subfigure[Adjusted Model Output vs DFT Potential Energy.]{
        \includegraphics[width=0.42\linewidth]{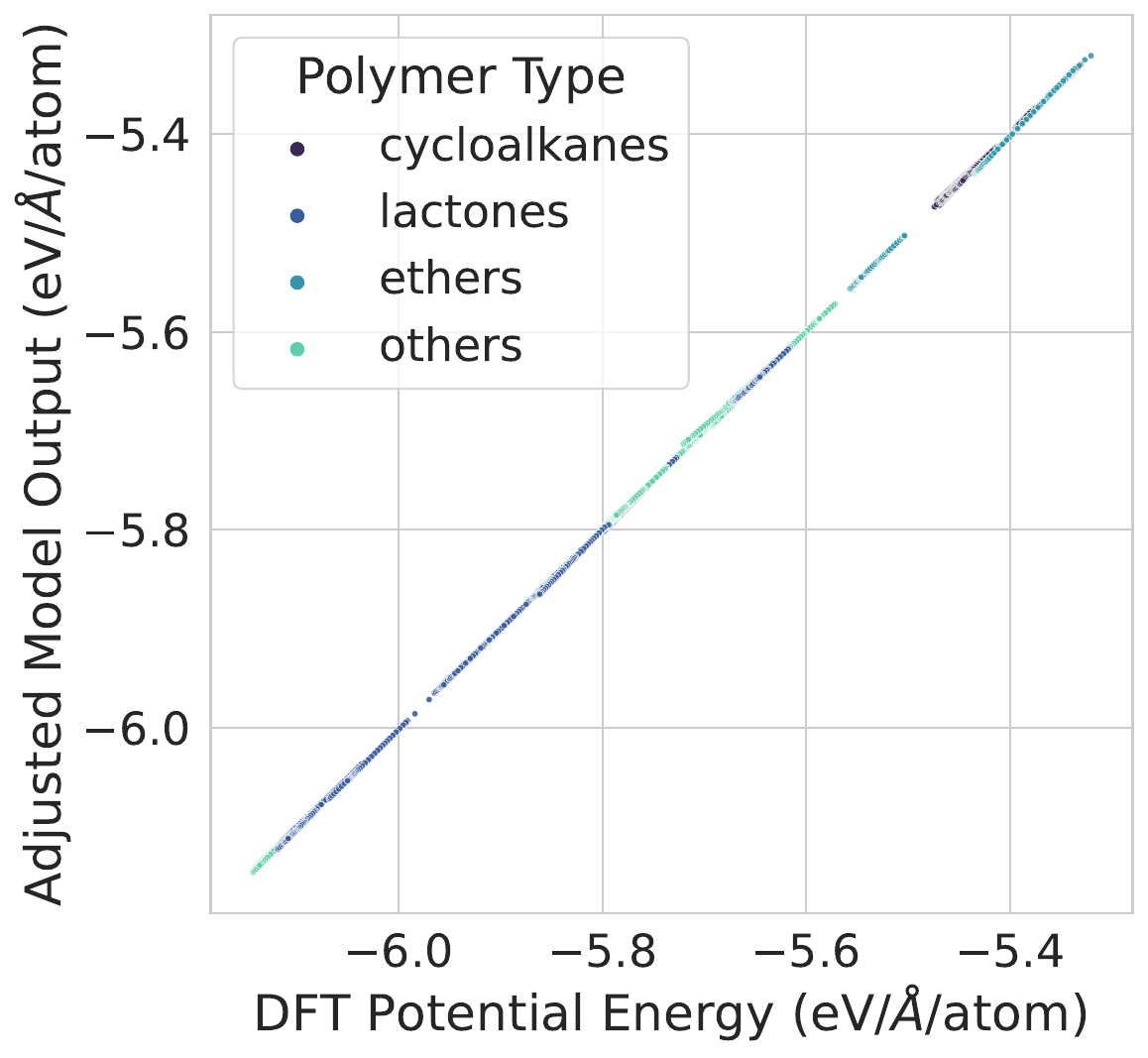}
        \label{fig:exp:adjusted_energy}
    }
    \caption{\ours's Energy Prediction Performance. Figure~\ref{fig:exp:unadjusted_energy}: \ours~model output plotted against DFT potential energy in eV/\AA/atom. Figure~\ref{fig:exp:adjusted_energy}: \ours~model predicts accurate potential energy after applying linear transforms for each polymer. }
    \label{fig:exp:energy_prediction}
\end{figure}

\paragraph*{Our model accurately predicts energy with linear transformation.}
Although our model does not directly optimize potential energy, it still captures the potential energy distributions for different polymers,  as shown in Figure~\ref{fig:exp:unadjusted_energy}. Averaged across different polymers, \ours~achieves an average of 0.0067 cosine distance between the model output and the DFT potential energy. After learning a linear regression on eight ground-truth conformations for each polymer, our model attains an average of 0.0006 eV/\AA/atom MAE between the transformed model energy prediction and the DFT potential energy and near 0 cosine distance. We can use this adjusted model output to predict accurate potential energies for polymers, as shown in ~\ref{fig:exp:adjusted_energy}.

%% file: exp_sim.tex
\subsection{MD Simulation Accuracy}
\label{sec:exp:sim}

\paragraph*{Model Training and Simulation Setup.}
We have developed a 6-layer Equivariant Transformer with an embedding dimension of 128 and 8 attention heads. During training, unless specified otherwise, we trained the model with 5-loop polymers and smaller ones. Polymers of size 6 and above were unseen during training and were used as test data. We used 5\% of the training data as the validation set and recorded the validation loss after every 500 training steps. The model has trained with the AdamW~\cite{you2019large} optimizer for 3 epochs with an initial learning rate of \(0.0001\), and the learning rate was reduced by a factor of \(0.8\) whenever the validation loss increased for 30 consecutive validation steps.

For testing the model's MD simulation performance, we take the 15-loop
polymerization of 4 example polymers from Table~\ref{table:data}, CK6,
OTH4, LAC2, and LAC5, and run MD simulations on them with \ours-generated
forcefields for a maximum of 600K steps. The number of atoms for these
15-loop polymers are, respectively: 360, 360, 180, and 240. The maximum loop size is seen in the training data
of \ours~ is 5-loop. Therefore, \ours~ is required to perform accurate MD
simulations on unseen polymers, testing both its simulation and
generalization ability.

For simulation, we have integrated our method into the ASE~\cite{larsen2017atomic} simulation environment. A diagram of the simulation is shown in Figure~\ref{fig:model:archi}. We use Nosé–Hoover thermostat for all simulations. The velocities for molecules are initialized with Maxwell-Boltzmann distribution to simulate the desired temperature environment. Unless otherwise specified, we use a temperature of 300 Kelvin and a timestep of 0.5fs. 


\paragraph*{Results.} 
 Figure~\ref{fig:large_polymer} (a)-(d) visualize the comparison results between \ours~ predicted forces and DFT reference data. 
For all 15 loop polymers, \ours~ obtains highly accurate forces to DFT calculations, with 0.01 eV/\(\AA\) MAE in energy and close to zero cosine distance. 
Hence, \ours~ can perform MD simulations on multiple types of polymers with uniformly high correlation with DFT references. The test 15-loop polymers contain up to 360 atoms, unseen in the training data.  
This shows that \ours~ can extrapolate well to large unseen polymers with known monomers. Furthermore, in practice, \ours~ only needs training data from small polymers, which are cheap to generate \emph{ab initio} data and thus greatly reduce the cost of DFT data generation.


In Figure~\ref{fig:large_polymer} (e) to (h), the Root Mean Square Deviation (RMSD) and potential energy graphs are shown. RMSD curve measures how much the polymer structure deviates from the initial conformation, and the potential energy graph indicates the energy state of the simulated polymers. Both RMSD and potential energy curves indicate that the polymer in simulation converges to a stable distribution around the equilibrium. Furthermore, \ours~ can explore different possible conformation states of the polymers, as illustrated in Figure~\ref{fig:large_polymer} (i) to (l).

\paragraph*{Simulation Efficiency.}
The \ours~model was trained on all polymers with loop size 5 and below using two A5000 GPUs, which took approximately 30 hours. However, this is a one-time cost, as the trained model can serve as a unified model for all future simulations of different polymers. 

For 15-loop polymers, the model predicts atomic forces for 20-30 steps per second on average. This means that the simulations for 600K steps, as mentioned in the previous section, can be completed in 6-8 hours.  
In contrast, the DFT model takes hours on CPU parallelization and still 5-10 minutes on GPU parallelization for one single-step calculation --- we speed up DFT simulations by at least 6000 times. Depending on the available GPU memory, the \ours~model can be parallelized to perform multiple simulations simultaneously. We leave the implementation of this acceleration for future work.

\hide{\paragraph*{When extrapolating to an unknown monomer structure \ours~ produces less accurate but still stable simulations. }

To further test the model, we constructed a chemically valid polymer, denoted by P0 and described by the SMILES string of [*]CC(O)CC(CO)CC(=O)CC[*].~\cite{doan2020machine} The distinct chemical feature of P0 is that it has two additional side-chain hydroxyl groups (-OH), which do not appear in the training data. We expect that our model, which works particularly well on the polymer models created from the monomers that are in the training data, may have different behaviors when seeing the new chemical features in P0.
}

%% file: related.tex
\section{Related Work}
\label{sec:related}

Machine learning forcefields have been attracting increasing attention, in order to achieve both high accuracy of \emph{ab initio} methods and the computation efficiency of empirical forcefields~\cite{ramprasad2017machine,unke2021machine,botu2017machine}. Traditional machine learning approaches rely on ``fingerprints'', pre-engineered interatomic features describing atomic environments. Similar to classical forcefields, these features describe pairwise radial or angular relations between atoms, and relatively simple machine learning algorithms such as kernel regression~\cite{chmiela2017machine,chmiela2018towards,chmiela2022accurate} and neural networks (NN)~\cite{unke2021machine,ramprasad2017machine,botu2017machine}  can transform these features to find the optimal statistical relations between the features and the potential energy and forces.
Particularly, neural networks have the universal approximation property~\cite{baker1998universal} that in theory can approximate arbitrarily complex functions, implying that with good enough features, the NN models can approximate \emph{ab initio} forces with high accuracy. Of course, in practice, optimality is difficult to achieve due to problems such as overfitting, model capacity, and the quality of feature engineering. Still, neural networks have been used in calculating forces for many different types of materials~\cite{kruglov2017energy,artrith2012high,behler2011atom,chen2017accurate,deringer2020general,hajibabaei2021machine}. 
Despite their success, traditional machine learning approaches rely on hand-crafted features that are not flexible enough to represent the complex quantum mechanical interactions between atoms. 
Imagine if the features are all based on pairwise distances between atoms. In such a case, no matter how complex the subsequent model is, the information it uses to make predictions would not extend beyond the pairwise relations defined in the features. This is not desirable if more complex quantum mechanics are involved.

The state-of-the-art approaches in machine learning forcefield take advantage of the
recent advances in Graph Neural Networks
(GNN). Instead of pre-defined fingerprint features, these algorithms model directly the interaction between atoms in the latent space. Properly designed with equivariant constraints, they have theoretically
universal approximation ability to equivariant and invariant physical
systems~\cite{,sannai2019universal}.
By treating molecules as graphs with 3D coordinates, GNN enables the learning of quantum mechanical interactions between atoms directly with equivariant operators, such as neural message passing~\cite{gilmer2017neural}, EGNN~\cite{satorras2021n}, SchNet~\cite{schutt2018schnet}, DIME~\cite{gasteiger2020directional}, GemNet~\cite{gasteiger2021gemnet}, and NequIP~\cite{batzner20223}, and TorchMDNet~\cite{tholke2022torchmd}. 
With stronger model capacity, they could achieve comparable accuracy with \emph{ab initio} forcefield. However, these models typically are only trained and evaluated on given conformations from single molecules~\cite{tholke2022torchmd,batzner20223,schutt2018schnet,gasteiger2021gemnet} (i.e., in-distribution data),  and it is nontrivial to directly extend them to train in multi-molecule settings. 
Furthermore, existing machine learning methods are prone to instability on out-of-distribution data, limiting their performance in MD simulations where noises are accumulated for atomic positions. The instability of existing methods is empirically analyzed by \cite{fu2022forces}. 

Among existing machine learning forcefields, AGNI~\cite{ramprasad2017machine,botu2017machine} and sGDML~\cite{chmiela2017machine,chmiela2018towards} agree with us in focusing on forces optimization. Several different reasons are stated. \cite{chmiela2017machine} observes the amplification effect of the differentiation operator on noises. Combined with the inherently noisy nature of machine learning algorithms, \cite{chmiela2017machine} claims that models trained on potential energy would cause considerable noise in the forces predictions. 
\cite{ramprasad2017machine,botu2017machine} present a learning scheme where the \emph{local} forces of atoms are learned and used to simulate geometry optimizations and MD simulations, instead of the \emph{global} potential energies. 
Adding to their observations, we find empirical evidence that the potential energy and forces are competing in the optimization of machine learning algorithms and that the forces of atoms, being localized features that can be derived from local atomic environments, have the benefit of better generalization across different molecules. 

%% file: conclusion.tex
\section{Conclusion}

We introduce PolyGET, a novel polymer forcefields learning framework with Generalizable Equivariant Transformers. Focusing on force optimization exclusively, our training paradigm enables accurate molecular dynamics simulations and generalization across polymer families and unseen polymers. Our multi-molecule approach fosters robust, transferable knowledge of interatomic interactions, contrasting with existing forcefields that jointly optimize potential energy and forces. Our results on large polymer simulation datasets demonstrate PolyGET's accuracy, robustness, and speed, showcasing its potential for material science and industry applications.

In the future, we aim to extend PolyGET's training to polymers beyond carbon, hydrogen, and oxygen. Additionally, we will investigate the extent to which PolyGET performs on vastly different chemical families and explore efficient fine-tuning strategies to enhance the model's adaptability with minimal effort.